%% file: nonmonotone.tex
\documentclass{article}

\usepackage{microtype}
\usepackage{graphicx}
\usepackage{subfigure}
\usepackage{booktabs} 

\usepackage[hyphens]{url}
\usepackage[breaklinks]{hyperref}
\usepackage{booktabs}       
\usepackage{amsfonts}       
\usepackage{nicefrac}       
\usepackage{microtype}      
\usepackage[bottom]{footmisc}

\usepackage[normalem]{ulem}
\usepackage{graphicx}
\usepackage{subfigure}
\usepackage{amsmath}
\usepackage{amsthm}
\usepackage{mathtools}
\usepackage{comment}
\usepackage{multirow}
\usepackage{wrapfig}
\usepackage{xcolor}
\usepackage{array}
\usepackage{xspace}
  
\theoremstyle{definition}

\usepackage[accepted]{icml2019}

\newcommand{\angleend}{\ensuremath{\langle \text{end}\rangle}\xspace}

\def\equationautorefname~#1\null{Eq~#1\null}
\renewcommand{\sectionautorefname}{\S\kern-0.2em}
\renewcommand{\subsectionautorefname}{\S\kern-0.2em}
\renewcommand{\subsubsectionautorefname}{\S\kern-0.2em}
\makeatletter \newcommand{\ALC@uniqueautorefname}{line} \makeatother

\begin{document}

\twocolumn[
\icmltitle{Non-Monotonic Sequential Text Generation}




\begin{icmlauthorlist}
\icmlauthor{Sean Welleck}{nyu}
\icmlauthor{Kiant\'e Brantley}{umd}
\icmlauthor{Hal Daum\'e III}{umd,msr}
\icmlauthor{Kyunghyun Cho}{nyu,fair,cifar}

\end{icmlauthorlist}

\icmlaffiliation{nyu}{New York University}
\icmlaffiliation{umd}{University of Maryland, College Park}
\icmlaffiliation{cifar}{CIFAR Azrieli Global Scholar}
\icmlaffiliation{fair}{Facebook AI Research}
\icmlaffiliation{msr}{Microsoft Research}

\icmlcorrespondingauthor{Sean Welleck}{wellecks@nyu.edu}

\icmlkeywords{Machine Learning, ICML}

\vskip 0.3in
]



\printAffiliationsAndNotice{} 

\begin{abstract}
Standard sequential generation methods assume a pre-specified generation order, such as text generation methods which generate words from left to right. 
In this work, we propose a framework for training models of text generation that operate in non-monotonic orders; the model directly learns good orders, without any additional annotation. 
Our framework operates by generating a word at an arbitrary position, and then recursively generating words to its left and then words to its right, yielding a binary tree.
Learning is framed as imitation learning, including a coaching method which moves from imitating an oracle to reinforcing the policy's own preferences. 
Experimental results demonstrate that using the proposed method, it is possible to learn policies which generate text without pre-specifying a generation order, while achieving competitive performance with conventional left-to-right generation.
\end{abstract}

\section{Introduction} \label{sec:introduction} \input{introduction}
\section{Non-Monotonic Sequence Generation} \label{sec:generation} \input{generation}
\section{Learning for Non-Monotonic Generation} \label{sec:learning} \input{learning}
\section{Neural Net Policy Structure} \label{sec:policy} \input{policy}
\section{Experimental Results} \label{sec:experiments} \input{experiments}
\input{exp_unconditional}
\input{exp_conditional}

\section{Related Work} \label{sec:related} \input{related}
\section{Conclusion, Limitations \& Future Work} \label{sec:future} \input{future}

\section*{Acknowledgements}

We thank support by eBay, TenCent and NVIDIA. This work was partly supported by Samsung Advanced Institute of Technology (Next Generation Deep Learning: from pattern recognition to AI), Samsung Electronics (Improving Deep Learning using Latent Structure), Sloan Foundation Research Fellowship, NSF Louis Stokes Alliances for Minority Participation Bridge to Doctorate (\#1612736) and ACM SIGHPC/Intel Computational and Data Science Fellowship, and STCSM 17JC1404100/1. 

\bibliography{nonmonotone}
\bibliographystyle{icml2019}
\newpage
\section{Appendix} \label{sec:appendix} \input{appendix}

\end{document}

%% file: introduction.tex
Most sequence-generation models, from n-grams \citep{bahl1983maximum} to neural language models \citep{bengio2003neural} generate sequences in a purely left-to-right, monotonic order. This raises the question of whether alternative, non-monotonic orders are worth considering~\citep{ford2018importance}, especially given the success of ``easy first'' techniques in natural language tagging \citep{tsuruoka2005bidirectional}, parsing \citep{goldberg2010efficient}, and coreference \citep{stoyanov2012easy}, which allow a model to effectively learn their own ordering. In investigating this question, we are solely interested in considering non-monotonic generation that does not rely on external supervision, such as parse trees~\citep{eriguchi2017learning,aharoni2017towards}.

\begin{figure}[t]
\centering \vspace{-1.0em}
\includegraphics[width=0.8\columnwidth]{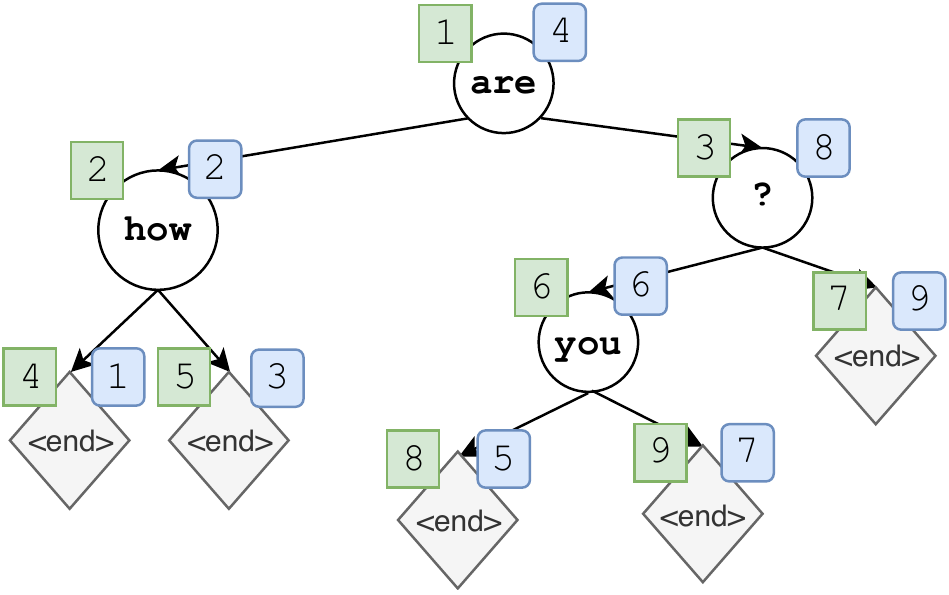}
\vspace{-5mm}\caption{
\label{fig:traversal-example}
A sequence, ``how are you ?'', generated by the proposed approach trained on utterances from a dialogue dataset. 
The model first generated the word ``are'' and then recursively generated left and right subtrees (``how'' and ``you ?'', respectively) of this word.
At each production step, the model may either generate a token, or an \angleend token, which indicates that this subtree is complete.
The full generation is performed in a level-order traversal, and the output is read off from an in-order traversal.
The numbers in green squares denote generation order (level-order); those in rounded blue squares denote location in the final sequence (in-order).
}
\vspace{-5mm}
\end{figure}

In this paper, we propose a framework for training sequential text generation models which learn a generation order without having to specifying an order in advance (\autoref{sec:generation}).
An example generation from our model is shown in \autoref{fig:traversal-example}.
We frame the learning problem as an \emph{imitation learning} problem, in which we aim to learn a generation policy that mimics the actions of an oracle generation policy (\autoref{sec:learning}).
Because the tree structure is unknown, the oracle policy cannot know the exact correct actions to take; to remedy this we propose a method called \emph{annealed coaching} which can yield a policy with learned generation orders, by gradually moving from imitating a maximum entropy oracle to reinforcing the policy's own preferences. 
Experimental results demonstrate that using the proposed framework, it is possible to learn policies which generate text without pre-specifying a generation order, achieving easy first-style behavior. The policies achieve performance metrics that are competitive with or superior to conventional left-to-right generation in language modeling, word reordering, and machine translation (\autoref{sec:experiments}).\footnote{Code and trained models available at \url{https://github.com/wellecks/nonmonotonic_text}.}


%% file: generation.tex
Formally, we consider the problem of sequentially generating a sequence of discrete tokens $Y=(w_1, \ldots, w_N)$, such as a natural language sentence, where $w_i \in V$, a finite vocabulary.
Let $\tilde V = V \cup \{ \angleend \}$.

Unlike conventional approaches with a fixed generation order, 
often left-to-right (or right-to-left), our goal is to build a sequence generator that generates these tokens in an order automatically determined by the sequence generator, without any extra annotation nor supervision of what might be a good order. We propose a method which does so by generating a word at an arbitrary position, then recursively generating words to its left and words to its right, yielding a binary tree like that shown in \autoref{fig:traversal-example}. 

We view the generation process as deterministically navigating a state space $\mathcal{S} = \tilde V^\star$ where a state $s \in \mathcal{S}$ corresponds to a sequence of tokens from $\tilde V$. We interpret this sequence of tokens as a top-down traversal of a binary tree, where \angleend terminates a subtree. The initial state $s_0$ is the empty sequence. For example, in \autoref{fig:traversal-example}, $s_1 = \langle \text{are} \rangle$, $s_2 = \langle \text{are}, \text{how} \rangle$, \dots, $s_4 = \langle \text{are}, \text{how}, \text{?}, \angleend \rangle$. An action $a$ is an element of $\tilde V$ which is deterministically appended to the state. Terminal states are those for which all subtrees have been \angleend'ed.
If a terminal state $s_T$ is reached, we have that $T = 2N+1$, where $N$ is the number of words (non-\angleend tokens) in the tree. We use $\tau(t)$ to denote the level-order traversal index of the $t$-th node in an in-order traversal of a tree, so that $\langle a_{\tau(1)}, \ldots, a_{\tau(T)} \rangle$ corresponds to the sequence of discrete tokens generated. The final sequence returned is this, postprocessed by removing all $\left< \text{end} \right>$'s. In \autoref{fig:traversal-example}, $\tau$ maps from the numbers in the blue squares to those in the green squares.

A policy $\pi$ is a (possibly) stochastic mapping from states to actions, and we denote the probability of an action $a \in \tilde V$ given a state $s$ as $\pi(a | s)$.
A policy $\pi$'s behavior decides which and whether words appear before and after the token of the parent node. 
Typically there are many unique binary trees with an in-order traversal equal to a sequence $Y$. Each of these trees has a different level-order traversal, thus the policy is capable of choosing from many different generation orders for $Y$, rather than a single predefined order.
Note that left-to-right generation can be recovered if $\pi(\angleend | s_t) = 1$ if and only if $t$ is odd (or non-zero and even for right-to-left generation).


%% file: learning.tex
Learning in our non-monotonic sequence generation model (\autoref{sec:generation}) amounts to inferring a policy $\pi$ from data. We first consider the \emph{unconditional} generation problem (akin to language modeling) in which the data consists simply of sequences $Y$ to be generated. Subsequently (\autoref{sec:conditional}) we consider the conditional case in which we wish to learn a mapping from inputs $X$ to output sequences $Y$.

This learning problem is challenging because the sequences $Y$ alone only tell us what the final output sequences of words should be, but not what tree(s) should be used to get there.
In left-to-right generation, the observed sequence $Y$ fully determines the sequence of actions to take.
In our case, however, the tree structure is effectively a latent variable, which will be determined by the policy itself.
This prevents us from using conventional supervised learning for training the parameterized policy.
On the other hand, at training time, we do know \emph{which words} should eventually appear, and their order; this substantially constrains the search space that needs to be explored, suggesting learning-to-search \citep{daume2009search} and imitation learning \citep{ross2011reduction,ross2014reinforcement} as a learning strategy.%
\footnote{One could consider applying reinforcement learning to this problem. This would ignore the fact that at training time we \emph{know} which words will appear, reducing the size of the feasible search space from $O(|V|^T)$ to $O(|X|^T)$, a huge savings. Furthermore, even with a fixed generation order, RL has proven to be difficult without partially relying on supervised learning~\citep{ranzato2015sequence,bahdanau2015task,bahdanau2016actor}.}

\begin{figure}[t]
    \centering
    \includegraphics[width=\columnwidth]{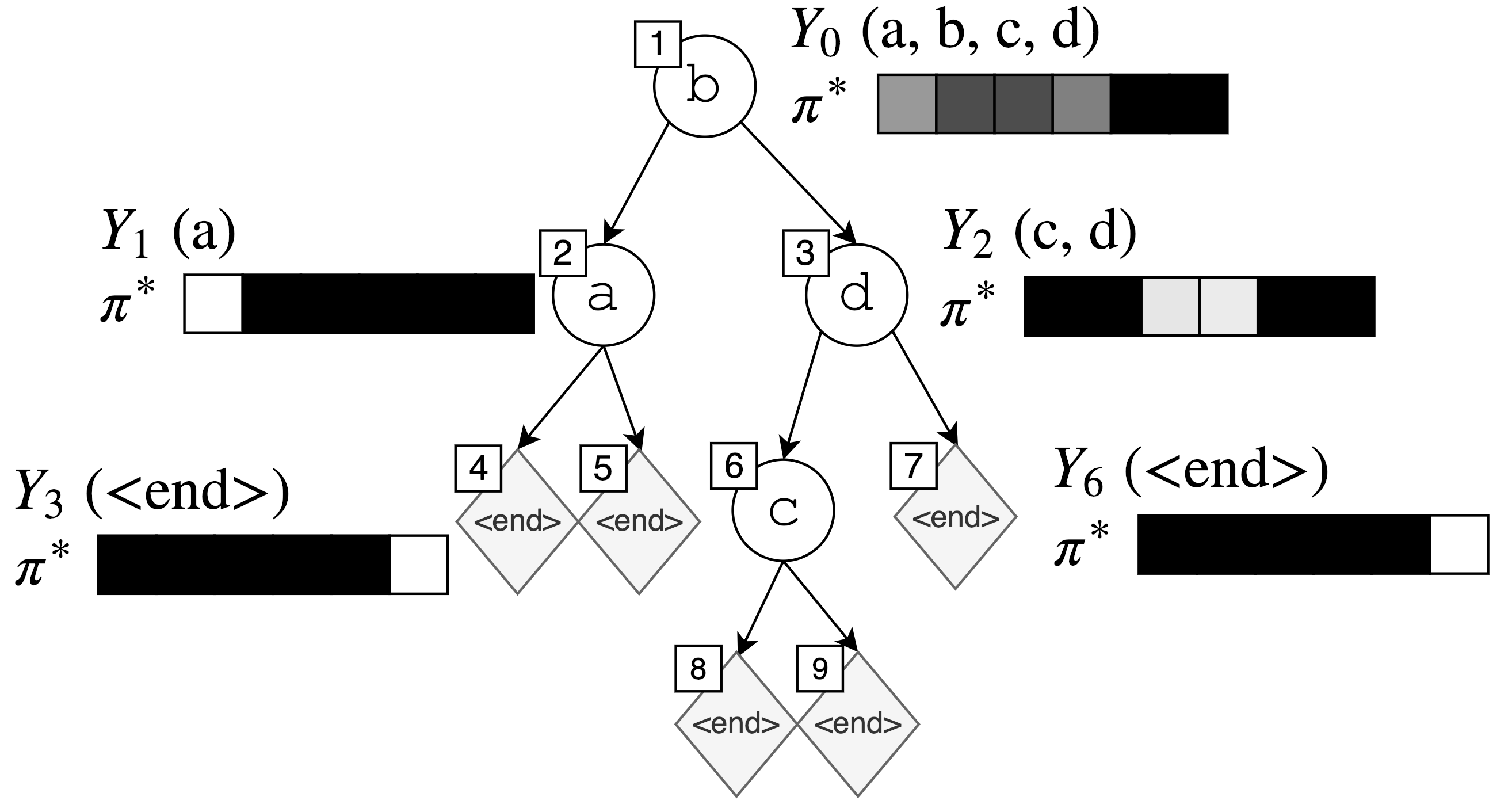}
    \vspace{-8mm}
    \caption{A sampled tree for the sentence ``a b c d'' with an action space $\tilde V=(\text{a,b,c,d,e,\angleend})$, showing an oracle's distribution $\pi^*$ and consecutive subsequences (``valid actions'') $Y_t$ for $t\in\{0,1,2,3,6\}$. Each oracle distribution is depicted as 6 boxes showing $\pi^*(a_{t+1}|s_t)$ (lighter = higher probability). 
    After \texttt{b} is sampled at the root, two empty left and right child nodes are created, associated with valid actions (a) and (c, d), respectively. 
    Here, $\pi^*$ only assigns positive probability to tokens in $Y_t$. 
    \label{fig:example_oracle}
    } 
    \vspace{-3mm}
\end{figure}

\textbf{The key idea} in our imitation learning framework is that at the first step, an oracle policy's action is to produce \emph{any} word $w$ that appears anywhere in $Y$.
Once picked, in a quicksort-esque manner, all words to the left of $w$ in $Y$ are generated recursively on the left (following the same procedure), and all words to the right of $w$ in $Y$ are generated recursively on the right.
(See \autoref{fig:example_oracle} for an example.)
Because the oracle is \emph{non-deterministic} (many ``correct'' actions are available at any given time), we inform this oracle policy with the current learned policy, encouraging it to favor actions that are preferred by the current policy, inspired by work in direct loss minimization \citep{hazan2010direct} and related techniques \citep{chiang2012hope,he2012imitation}.

\subsection{Background: Learning to Search}

\newcommand{\piin}{{\pi^{\text{in}}}}
\newcommand{\piout}{{\pi^{\text{out}}}}

In learning-to-search-style algorithms, we aim to learn a policy $\pi$ that mimics an oracle (or ``reference'') policy $\pi^*$.
To do so, we define a \emph{roll-in} policy $\piin$ and \emph{roll-out} policy $\piout$.
We then repeatedly draw states $s$ according to the state distribution induced by $\piin$, and compute cost-to-go under $\piout$, for all possible actions $a$ at that state.
The learned policy $\pi$ is then trained to choose actions to minimize this cost-to-go estimate.

Formally, denote the uniform distribution over $\{1, \dots, T\}$ as $U[T]$ and denote by $d_\pi^t$ the distribution of states induced by running $\pi$ for $t$-many steps.
Denote by $\mathcal{C}(\pi; \piout, s)$ a scalar cost measuring the loss incurred by $\pi$ against the cost-to-go estimates under $\piout$ (for instance, $\mathcal{C}$ may measure the squared error between the vector $\pi(\cdot|s)$ and the cost-to-go estimates).
Then, the quantity being optimized is:
\begin{align}
\label{eq:ls-loss}
    \mathbb{E}_{Y \sim D}
    \mathbb{E}_{t \sim U[2|Y|+1]}
    \mathbb{E}_{s_t \sim d_\piin^t} 
    \left[
    \mathcal{C}(\pi; \piout, s_t)
    \right]
\end{align}
Here, $\piin$ and $\piout$ can use information not available at test-time (e.g., the ground-truth $Y$). Learning consists of finding a policy which only has access to states $s_t$ but performs as well or better than $\pi^*$.
By varying the choice of $\piin$, $\piout$, and $\mathcal{C}$, one obtains different variants of learning-to-search algorithms, such as 
DAgger~\citep{ross2011reduction}, AggreVaTe~\citep{ross2014reinforcement} or LOLS~\citep{chang2015learning}.

In the remainder of this section, we describe the cost function we use, a set of oracle policies and a set of roll-in policies, both of which are specifically designed for the proposed problem of non-monotonic sequential generation of a sequence. These sets of policies are empirically evaluated later in the experiments (\autoref{sec:experiments}).

\subsection{Cost Measurement}

There are many ways to measure the prediction cost $\mathcal{C}(\pi; \piout, s)$; arguably the most common is squared error between cost-predictions by $\pi$ and observed costs obtained by $\piout$ at the state $s$.
However, recent work has found that, especially when dealing with recurrent neural network policies (which we will use; see \autoref{sec:policy}), using a cost function more analogous to a cross-entropy loss can be preferred \cite{leblond2018searnn,cheng2018fast,welleck2018loss}.
In particular, we use a KL-divergence type loss, measuring the difference between the action distribution produced by $\pi$ and the action distribution preferred by $\piout$.
\begin{align}
\label{eq:dkl}
    \mathcal{C}(\pi; \piout, s) &= D_{\text{KL}}\left(\piout(\cdot|s)~||~\pi(\cdot|s)\right)\\
    &= \sum_{a\in \tilde V}\piout(a|s)\log \pi(a|s)+\textit{const.}\nonumber
\end{align}
%
%
Our approach estimates the loss in Eq.~\eqref{eq:ls-loss} by first sampling one training sequence, running the roll-in policy for $t$ steps, and computing the KL divergence \eqref{eq:dkl} at that state using $\pi^*$ as $\pi^{\text{out}}$. Learning corresponds to minimizing this KL divergence iteratively with respect to the parameters of $\pi$.

\subsection{Roll-In Policies}

The \emph{roll-in} policy determines the state distribution over which the learned policy $\pi$ is to be trained.
In most formal analyses, the roll-in policy is a stochastic mixture of the learned policy $\pi$ and the oracle policy $\pi^*$, ensuring that $\pi$ is eventually trained on its own state distribution~\citep{daume2009search,ross2011reduction,ross2014reinforcement,chang2015learning}.
Despite this, \emph{experimentally}, it has often been found that simply using the oracle's state distribution is optimal \citep{ranzato2015sequence,leblond2018searnn}. This is likely because the noise incurred early on in learning by using $\pi$'s state distribution is not overcome by the benefit of matching state distributions, especially when the the policy class is sufficiently high capacity so as to be nearly realizable on the training data \citep{leblond2018searnn}.
In preliminary experiments, we observed the same is true in our setting: simply rolling in according to the oracle policy (\autoref{sec:oracle}) yielded the best results experimentally.
Therefore, despite the fact that this can lead to inconsistency in the learned model \citep{chang2015learning}, all experiments are with oracle roll-ins.

\subsection{Oracle Policies} \label{sec:oracle}

In this section we formalize the oracle policies that we consider.
To simplify the discussion (we assume that the roll-in distribution is the oracle), we only need to define an oracle policy that takes actions on states it, itself, visits. 
All the oracles we consider have access to the ground truth output $Y$, and the current state $s$.
We interpret the state $s$ as a partial binary tree and a ``current node'' in that binary tree where the next prediction will go.
It is easiest to consider the behavior of the oracle as a top-down, level-order traversal of the tree, where in each state it maintains a sequence of ``possible tokens'' at that state.
An oracle policy $\pi^*(\cdot | s_t)$ is defined with respect to $Y_t$, a consecutive subsequence of $Y$. At $s_0 = \langle \rangle$, $\pi^*$ uses the full $Y_0=Y$. This is subdivided as the tree is descended.
At each state $s_t$, $Y_t$ contains ``valid actions''; labeling the current node with \textit{any} token from $Y_t$ keeps the generation leading to $Y$. 
For instance, in \autoref{fig:example_oracle}, after sampling \textit{b} for the root, the valid actions $(a, b, c, d)$ are split into $(a)$ for the left child and $(c, d)$ for the right child.

Given the consecutive subsequence $Y_t=(w'_1, \ldots, w'_{N'})$, an oracle policy is defined as:
\begin{align}
\label{eq:generic-oracle}
    \pi^*(a | s_t) =
    \begin{cases}
      1 & \text{if } a=\angleend \text{ and } Y_t = \langle\rangle \\
      p_a &\text{if } a \in Y_t \\
      0 &\text{otherwise}
    \end{cases}
\end{align}
where the $p_a$s are arbitrary such that $\sum_{a \in Y} p_a = 1$. An oracle policy places positive probability only on valid actions, and forces an \angleend output if there are no more words to produce. 
This is guaranteed to \emph{always} generate $Y$, regardless of how the random coin flips come up.

When an action $a$ is chosen, at $s_t$, this ``splits'' the sub-sequence $Y_t = (w'_1, \dots, w'_{N'})$ into left and right sub-sequences, $\overleftarrow{Y}_t=(w'_1, \ldots, w'_{i-1})$ and $\overrightarrow{Y}_t=(w'_{i+1}, \ldots, w_N)$,
where $i$ is the index of $a$ in $Y_t$. (This split may not be unique due to duplicated words in $Y_t$, in which case we choose a valid split arbitrarily.) 
These are ``passed'' to the left and right child nodes, respectively. 

There are many possible oracle policies, and each of them is characterized by how $p_a$ in Eq.~\eqref{eq:generic-oracle} is defined. Specifically, we propose three variants.

\paragraph{Uniform Oracle.}
Motivated by \citet{welleck2018loss} who applied learning-to-search to the problem of multiset prediction, we design a uniform oracle $\pi^*_{\text{uniform}}$.
This oracle treats all possible generation orders that lead to the target sequence $Y$ as equally likely, without preferring any specific set of orders.
Formally, $\pi^*_{\text{uniform}}$ gives uniform probabilities $p_a = 1/n$ for all words in $Y_t$ where $n$ is the number of unique words in $Y_t$. (\citet{daume09unsearn} used a similar oracle for unsupervised structured prediction, which has a similar non-deterministic oracle complication.)

\paragraph{Coaching Oracle.}
An issue with the uniform oracle is that it does not prefer any specific set of generation orders, making it difficult for a parameterized policy to imitate. This gap has been noticed as a factor behind the difficulty in learning-to-search by \citet{he2012imitation}, who propose the idea of coaching. In coaching, the oracle takes into account the preference of a parameterized policy in order to facilitate its learning. Motivated by this, we design a coaching oracle as the product of the uniform oracle and current policy $\pi$:
\begin{align}
\label{eq:coaching-oracle}
    \pi^*_{\text{coaching}}(a|s) \propto \pi^*_{\text{uniform}}(a|s) ~ \pi(a|s)
\end{align}
This coaching oracle ensures that no invalid action is assigned any probability, while preferring actions that are preferred by the current parameterized policy, reinforcing the selection by the current policy if it is valid.

\paragraph{Annealed Coaching Oracle.}
The multiplicative nature of the coaching oracle gives rise to an issue, especially in the early stage of learning, as it does not encourage learning to explore a diverse set of generation orders. We thus design a mixture of the uniform and coaching policies, which we refer to as an annealed coaching oracle:
\begin{align}
    \label{eq:annealed-oracle}
    \pi_{\text{annealed}}^*(a|s)
    &= 
      \beta \pi^*_{\text{\scalebox{.92}[1.0]{uniform}}} (a|s)
      + 
      (1-\beta) \pi^*_{\text{\scalebox{.92}[1.0]{coaching}}}(a|s)
\end{align}
We anneal $\beta$ from $1$ to $0$ over learning, on a linear schedule.

\paragraph{Deterministic Left-to-Right Oracle.}
In addition to the proposed oracle policies above, we also experiment with a deterministic oracle that corresponds to generating the target sequence from left to right: $\pi^*_{\text{left-right}}$ always selects the first un-produced word as the correct action, with probability $1$.
When both roll-in and oracle policies are set to the left-to-right oracle $\pi^*_{\text{left-right}}$, the proposed approach recovers to maximum likelihood learning of an autoregressive sequence model, which is {\it de facto} standard in neural sequence modeling. In other words, supervised learning of an autoregressive sequence model is a special case of the proposed approach.


%% file: policy.tex
We use a neural network to implement the proposed binary tree generating policy, as it has been shown to encode a variable-sized input and predict a structured output effectively~\citep{cleeremans1989finite,forcada1997recursive,sutskever2014sequence,cho2014learning,tai2015improved,bronstein2017geometric,battaglia2018relational}. This neural network takes as input a partial binary tree, or equivalently a sequence of nodes in this partial tree by level-order traversal, and outputs a distribution over the action set $\tilde V$.

\paragraph{LSTM Policy.}
The first policy we consider is implemented as a recurrent network with long short-term memory (LSTM) units~\citep{hochreiter1997long} by considering the partial binary tree as a flat sequence of nodes in a level-order traversal $(a_1, \ldots, a_{t})$. The recurrent network encodes the sequence into a vector $h_t$ and computes a categorical distribution over the action set:
\begin{align}\label{eq:pi}
    \pi(a | s_t) \propto 
    \exp(u_a^\top h_t + b_a)
\end{align}
where $u_a$ and $b_a$ are weights and bias associated with $a$. 

This LSTM structure relies entirely on the linearization of a partial binary tree, and minimally takes advantage of the actual tree structure \emph{or} the surface order. It may be possible to exploit the tree structure more thoroughly using a recurrent architecture that is designed to encode a tree~\citep{zhang2015top,alvarez2016tree,dyer2015transition,bowman2016fast}, which we leave for future investigation. We did experiment with additionally conditioning $\pi$'s action distribution on the \emph{parent} of the current node in the tree, but preliminary experiments did not show gains.

\paragraph{Transformer Policy.} We additionally implement a policy using a Transformer \cite{vaswani2017attention}. The level-order sequence $a_1,...,a_t$ is again summarized by a vector $h_t$, here computed using a multi-head attention mechanism. 
As in the LSTM policy, the vector $h_t$ is used to compute a categorical distribution over the action set \eqref{eq:pi}.

\paragraph{Auxiliary \angleend Prediction.} We also consider separating the action prediction into token ($a_i\in \mathcal{V}$) prediction and \angleend prediction. The policy under this view consists of a categorical distribution over tokens \eqref{eq:pi} as well as an \angleend predictor which parameterizes a Bernoulli distribution, $\pi_{\text{end}}(\angleend | s_t) \propto 
    \sigma(u_e^\top h_t + b_e),$
where $\pi_{\text{end}}(\angleend =1| s_t)$ means $a_t$ is \angleend, and $a_t$ is determined by $\pi$ according to \eqref{eq:pi} otherwise. At test time, we threshold the predicted \angleend probability at a threshold $\tau$. In our experiments, we only use this approach with the Transformer policy (\autoref{sec:mt}).

\subsection{Conditional Sentence Generation} \label{sec:conditional}

An advantage of using a neural network to implement the proposed policy is that it can be easily conditioned on an extra context. It allows us to build a conditional non-monotonic sequence generator that can for instance be used for machine translation, image caption generation, speech recognition and generally multimedia description generation~\citep{cho2015describing}. 
To do so, we assume that a conditioning input (e.g. an image or sentence) $X$ can be represented as a set of $d_{\text{enc}}$-dimensional context vectors, obtained with a learned \textit{encoder} function $f^{\text{enc}}(X)$ whose parameters are learned jointly with the policy's.

For word-reordering experiments \autoref{sec:reorder}, the encoder outputs a single vector which is used to initialize the LSTM policy's state $h_0$. In the machine translation experiments \autoref{sec:mt}, the Transformer encoder outputs $|X|$ vectors, $H\in \mathbb{R}^{|X|\times d_{\text{enc}}}$, which are used as input to a decoder (i.e. policy) attention function; see \citep{vaswani2017attention} for further details.


%% file: experiments.tex
In this section we experiment with our non-monotone sequence generation model across four tasks.
The first two are \emph{unconditional} generation tasks: language modeling (\autoref{sec:lm}) and out-of-order sentence completion (\autoref{sec:completion}).
Our analysis in these tasks is primarily qualitative: we seek to understand what the non-monotone policy is learning and how it compares to a left-to-right model.
The second two tasks are \emph{conditional} generation tasks, which generate output sequences based on some given input sequence: word reordering (\autoref{sec:reorder}) and machine translation (\autoref{sec:mt}).


%% file: exp_unconditional.tex
\subsection{Language Modeling} \label{sec:lm}

We begin by considering generating samples from our model, trained as a language model.
Our goal in this section is to qualitatively understand what our model has learned.
It would be natural also to evaluate our model according to a score like perplexity.
Unfortunately, unlike conventional autoregressive language models, it is intractable to compute the probability of a given sequence in the non-monotonic generation setting, as it requires us to marginalize out all possible binary trees that lead to the sequence.

\label{ssec:unc_setup}
\paragraph{Dataset.}
We use a dataset derived from the Persona-Chat \citep{zhang2018personalizing} dialogue dataset, which consists of multi-turn dialogues between two agents. 
Our dataset here consists of all unique persona sentences and utterances in Persona-Chat.
We derive the examples from the same train, validation, and test splits as Persona-Chat, resulting in 133,176 train, 16,181 validation, and 15,608 test examples. Sentences are tokenized by splitting on spaces and punctuation. The training set has a vocabulary size of 20,090 and an average of 12.0 tokens per example.

\paragraph{Model.} We use a uni-directional LSTM that has 2 layers of 1024 LSTM units. 
See Appendix A.2 for more details.

\begin{table}[t]
  \centering
    \footnotesize
  \begin{tabular}{@{}lccp{1.0cm}p{0.8cm}p{0.8cm}@{}}
	\toprule
	\textbf{Oracle} & \textbf{\%Novel} & \textbf{\%Unique} & \textbf{Avg. Tokens} & \textbf{Avg. Span} & \textbf{\textsc{Bleu}} \\ 
	\midrule
	left-right & 17.8  & 97.0 &  11.9 & 1.0 & 47.0\\
	uniform   & 98.3 & 99.9 &  13.0 & 1.43 & 40.0\\
	annealed  & 93.1 & 98.2 &  10.6 & 1.31 & 56.2\\
	\midrule
	Validation     & 97.0 & 100 & 12.1 & - & -\\
	\bottomrule
\end{tabular}%
\vspace{-3mm}
\caption{\label{tbl:unc-sample-stats}
Statistics computed over 10,000 sampled sentences (in-order traversals of sampled trees with $\langle end\rangle$ tokens removed) for policies trained on Persona-Chat. A sample is novel when it is not in the training set. Percent unique is the cardinality of the set of sampled sentences divided by the number of sampled sentences.}
\vspace{-2mm}
\end{table}
\begin{table}[t]
  \footnotesize
  \centering
\begin{tabular}{@{}l@{~}m{5cm}l@{}}
	\toprule
\textbf{$\pi^*$} & \textbf{Samples} \\
  \midrule
  \parbox[t]{2mm}{\multirow{6}{*}{\rotatebox[origin=c]{90}{left-right}}}
& $\circ$ hey there , i should be ! & \hspace{-0.3cm}\multirow{5}{*}{\includegraphics[height=60px]{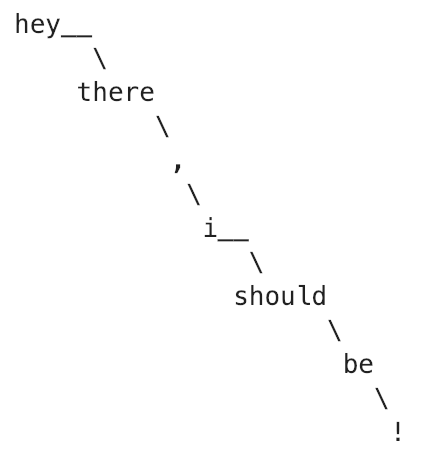}} \\
& $\circ$ not much fun . what are you doing ?\\
& $\circ$ not . not sure if you .\\
& $\circ$ i love to always get my nails done .\\
& $\circ$ sure , i can see your eye underwater\\&\qquad while riding a footwork .\\
\midrule
  \parbox[t]{2mm}{\multirow{7}{*}{\rotatebox[origin=c]{90}{uniform}}}
& $\circ$ i just got off work . &\hspace{-0.3cm}\multirow{5}{*}{\includegraphics[height=55px]{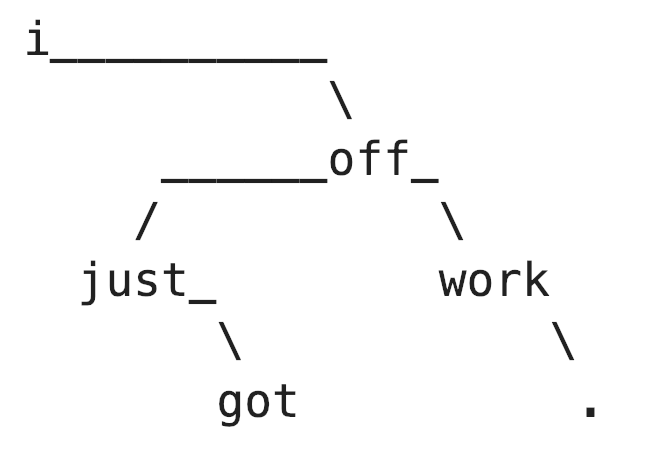}} \\
& $\circ$ yes but believe any karma , it is .\\
& $\circ$ i bet you are . i read most of good tvs \\&\qquad on that horror out . cool .\\
& $\circ$ sometimes , for only time i practice \\&\qquad professional baseball .\\
& $\circ$ i am rich , but i am a policeman .\\
\midrule
  \parbox[t]{2mm}{\multirow{7}{*}{\rotatebox[origin=c]{90}{annealed}}}
& $\circ$ i do , though . do you ? & \hspace{-0.3cm}\multirow{7}{*}{\includegraphics[height=60px]{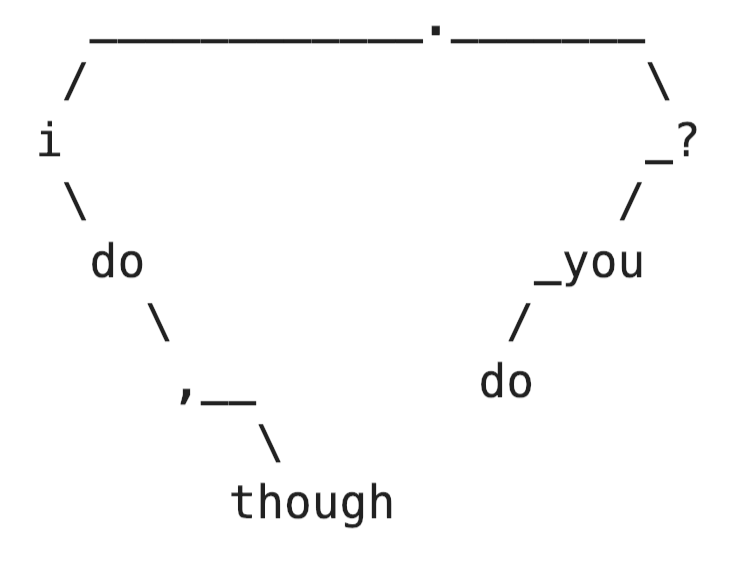}} \\
& $\circ$ i like iguanas . i have a snake . i wish\\&\qquad  i could win . you ?\\
& $\circ$ i am a homebody .\\
& $\circ$ i care sometimes . i also snowboard .\\
& $\circ$ i am doing okay . just relaxing , \\&\qquad and you ?\\
  \bottomrule
\end{tabular}
\vspace{-3mm}
\caption{\label{tbl:unc-samples}
Samples from unconditional generation policies trained on Persona-Chat for each training oracle. The first sample's underlying tree is shown. See Appendix A.2 for more samples.}
\vspace{-3mm}
\end{table}

\paragraph{Basic Statistics.}
We draw 10,000 samples from each trained policy (by varying the oracle) and analyze the results using the following metrics: percentage of novel sentences, percentage of unique, average number of tokens, average span size\footnote{The average span is the average number of children for non-leaf nodes excluding the special token \angleend, ranging from $1.0$ (chain, as induced by the left-right oracle) to $2.0$ (full binary tree).} and \textsc{Bleu} (\autoref{tbl:unc-sample-stats}). We use \textsc{Bleu}
to quantify the sample quality by computing the \textsc{Bleu} score of the samples using the validation set as reference, following \citet{yu2016seqgan} and \citet{zhu2018texygen}. In Appendix Table 6 we report additional scores.
We see that the non-monotonically trained policies generate many more novel sentences, and build trees that are bushy (span $\sim 1.3$), but not complete binary trees. The policy trained with the annealed oracle is most similar to the validation data according to \textsc{Bleu}.

\paragraph{Content Analysis.}
We investigate the content of the models in \autoref{tbl:unc-samples}, which shows samples from policies trained with different oracles. Each of the displayed samples are not a part of the training set. We provide additional samples organized by length in Appendix Tables 8 and 9, and samples showing the underlying trees that generated them in Appendix Figures 5-7. We additionally examined word frequencies and part-of-speech tag frequencies, finding that the samples from each policy typically follow the validation set's word and tag frequencies. 

\begin{figure}[t]
    \centering
    \includegraphics[width=\columnwidth]{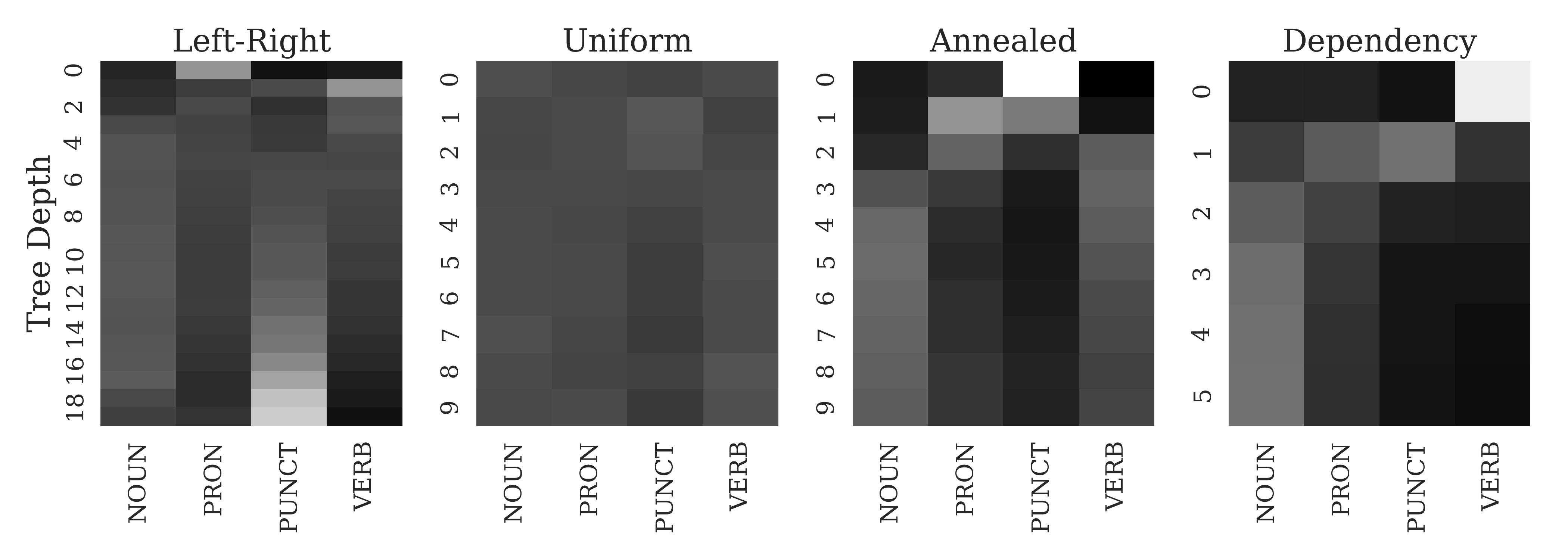}
\vspace{-8mm}
    \caption{    \label{fig:unc_pos_heatmap}
POS tag counts by tree-depth, computed by tagging 10,000 sampled sentences. Counts are normalized across each row (depth), then the marginal tag probabilities are subtracted. A light value means the probability of the tag occurring at that depth is higher than the prior probability of the tag occurring. 
    }
\vspace{-3mm}
\end{figure}

\paragraph{Generation Order.} 
We analyze the generation order of our various models by inspecting the part-of-speech (POS) tags each model tends to put at different tree depths  (i.e. number of edges from node to root). 
\autoref{fig:unc_pos_heatmap} shows POS counts by tree depth, normalized by the sum of counts at each depth (we only show the four most frequent POS categories). 
We also show POS counts for the validation set's dependency trees, obtained with an off-the-shelf parser.
Not surprisingly, policies trained with the uniform oracle tend to generate words with a variety of POS tags at each level. 
Policies trained with the annealed oracle on the other hand, learned to frequently generate punctuation at the root node, often either the sentence-final period or a comma, in an ``easy first'' style, since most sentences contain a period.
Furthermore, we see that the policy trained with the annealed oracle tends to generate a pronoun before a noun or a verb (tree depth 1), which is a pattern that policies trained with the left-right oracle also learn. Nouns typically appear in the middle of the policy trained with the annealed oracle's trees. Aside from verbs, the annealed policy's trees, which place punctuation and pronouns near the root and nouns deeper, follow a similar structure as the dependency trees. 
\subsection{Sentence Completion} \label{sec:completion}

A major weakness of the conventional autoregressive model, especially with unbounded context, is that it cannot be easily used to fill in missing parts of a sentence except at the end. This is especially true when the number of tokens per missing segment is not given in advance. Achieving this requires significant changes to both model architecture, learning and inference~\citep{berglund2015bidirectional}.

Our proposed approach, on the other hand, can naturally fill in missing segments in a sentence.
Using models trained as language models from the previous section (\autoref{sec:lm}), we can achieve this by initializing a binary tree with observed tokens in a way that they respect their relative positions.
For instance, the first example shown in Table \ref{tbl:unc-tree-complete} can be seen as the template ``\underline{~~~~~~~} favorite \underline{~~~~~~~} food \underline{~~~~~~~} ! \underline{~~~~~~~}'' with variable-length missing segments.
Generally, an initial tree with nodes $(w_i,\ldots,w_k)$ ensures that each $w_j$ appears in the completed sentence, and that $w_i$ appears at \textit{some} position to the left of $w_j$ in the completed sentence when $w_i$ is a left-descendant of $w_j$ (analogously for right-descendants).
\begin{table}[t]
  \footnotesize\centering
\begin{tabular}{@{}ll@{}}
	\toprule
	\textbf{Initial Tree} & \textbf{Samples} \\
\midrule
  \multirow{6}{*}{\includegraphics[width=2cm]{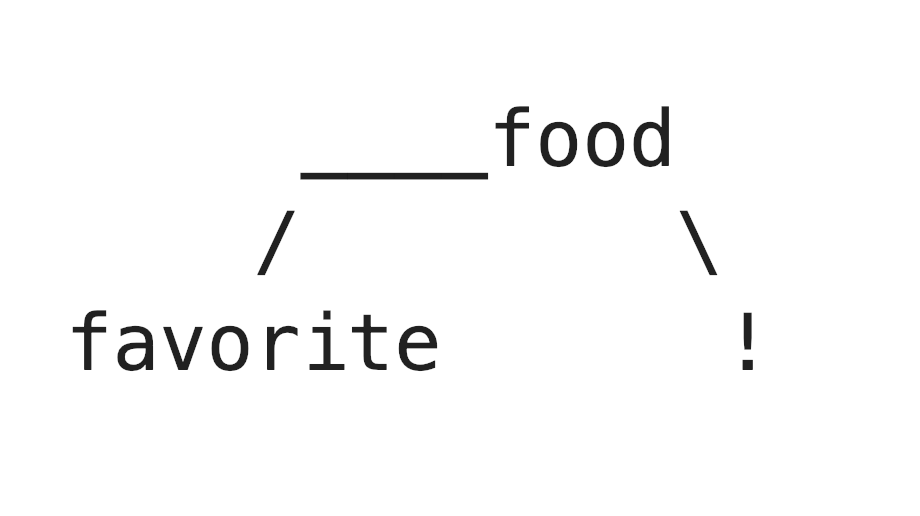}}
                              & $\circ$ lasagna is my \textbf{favorite food !} \\
                              & $\circ$ my \textbf{favorite} \textbf{food} is mac and cheese \textbf{!}\\
                              & $\circ$ what is your \textbf{favorite} \textbf{food} ? pizza , i love it \textbf{!}\\
                              & $\circ$ whats your \textbf{favorite} \textbf{food} ? mine is pizza \textbf{!}\\
                              & $\circ$ seafood is my \textbf{favorite} . and mexican \textbf{food} \textbf{!} \\&\qquad what is yours ?\\
\midrule
  \multirow{6}{*}{\includegraphics[width=2cm]{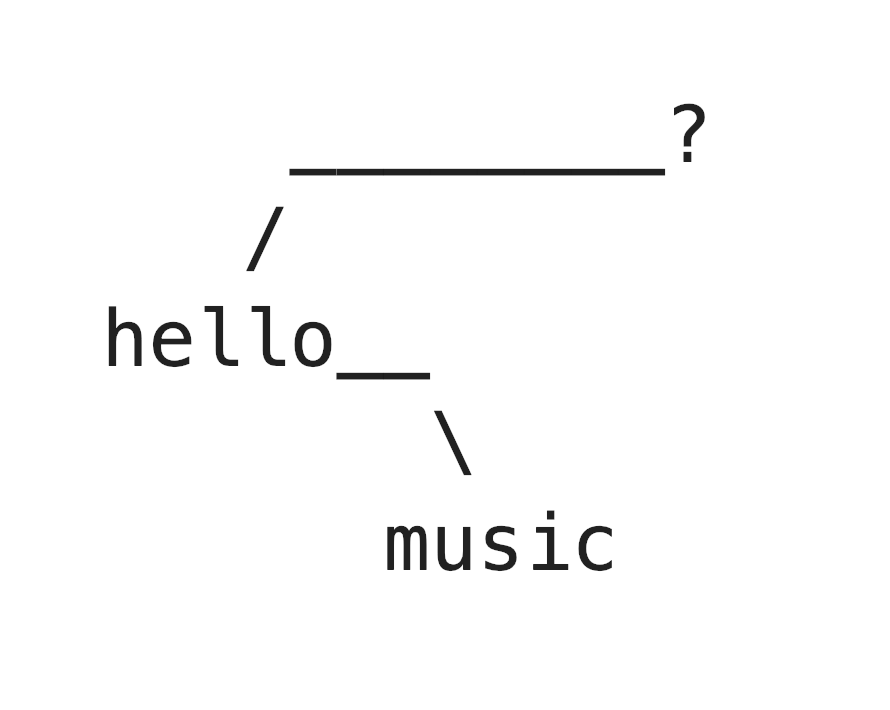}}
                              &$\circ$ \textbf{hello} ! i like classical \textbf{music} . do you \textbf{?} \\
                              &$\circ$ \textbf{hello} , do you enjoy playing \textbf{music} \textbf{?}\\
                              &$\circ$ \textbf{hello} just relaxing at home listening to \\&\qquad fine \textbf{music} . you \textbf{?}\\
                              &$\circ$ \textbf{hello} , do you like to listen to \textbf{music} \textbf{?}\\
                              &$\circ$ \textbf{hello} . what kind of \textbf{music} do you like \textbf{?}\\
\midrule
  \multirow{6}{*}{\includegraphics[width=2cm]{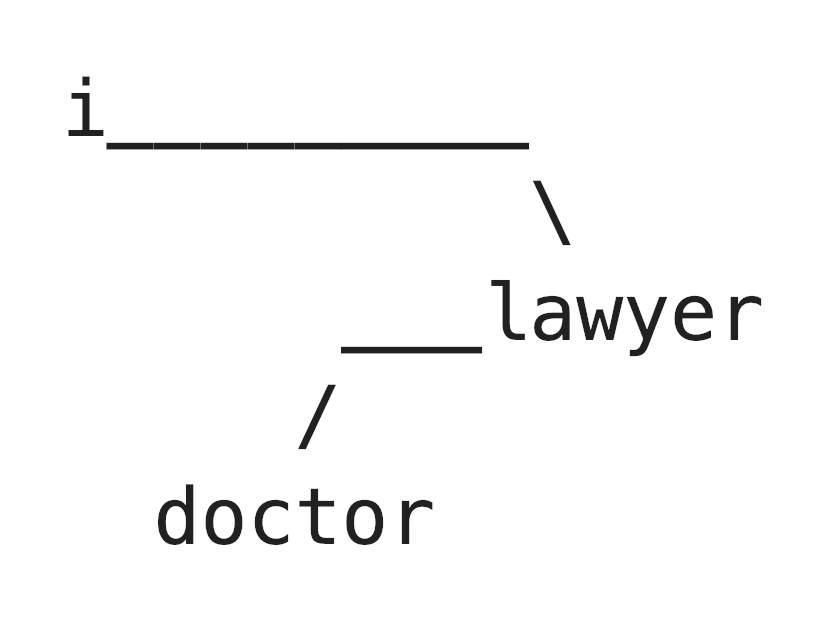}}
                              & $\circ$ \textbf{i} am a \textbf{doctor} or a \textbf{lawyer} . \\
                              & $\circ$ \textbf{i} would like to feed my \textbf{doctor} , i aspire  \\&\qquad to be a \textbf{lawyer} .\\
                              & $\circ$ \textbf{i} am a \textbf{doctor} \textbf{lawyer} . 4 years old .\\
                              & $\circ$ \textbf{i} was a \textbf{doctor} but went to a \textbf{lawyer} .\\
                              & $\circ$ \textbf{i} am a \textbf{doctor} since i want to be a \textbf{lawyer} .\\
\bottomrule
\end{tabular}
\caption{\label{tbl:unc-tree-complete}
Sentence completion samples from a policy trained on Persona-Chat with the uniform oracle. The left column shows the initial seed tree. In the sampled sentences, seed words are bold. 
}
\vspace{-3mm}
\end{table}

To quantify the completion quality, we first create a collection of initial trees by randomly sampling three words $(w_i, w_j, w_k)$ from each sentence $Y=(w_1,\ldots,w_T)$ from the Persona-Chat validation set of \autoref{ssec:unc_setup}. We then sample one completion for each initial tree and measure the \textsc{Bleu} of each sample using the validation set as reference as in \autoref{sec:lm}. According to \textsc{Bleu}, the policy trained with the annealed oracle sampled completions that were more similar to the validation data (\textsc{Bleu} 44.7) than completions from the policies trained with the uniform (\textsc{Bleu} 38.9) or left-to-right (\textsc{Bleu} 14.3) oracles.

In Table~\ref{tbl:unc-tree-complete}, we present some sample completions using the policy trained with the uniform oracle. The completions illustrate a property of the proposed non-monotonic generation that is not available in left-to-right generation.


%% file: exp_conditional.tex
\subsection{Word Reordering} \label{sec:reorder}

We first evaluate the proposed models for conditional generation on the Word Reordering task, also known as Bag Translation \citep{brown1990statistical} or Linearization \citep{schmaltz2016word}. In this task, a sentence $Y=(w_1,...,w_N)$ is given as an unordered collection $X=\{w_1,...,w_N\}$, and the task is to reconstruct $Y$ from $X$.
We assemble a dataset of $(X, Y)$ pairs using sentences $Y$ from the Persona-Chat sentence dataset of \autoref{ssec:unc_setup}.
In our approach, we do not explicitly force the policies trained with our non-monotonic oracles to produce a permutation of the input and instead let them learn this automatically.

\paragraph{Model.} For encoding each unordered input $x=\{w_1,...,w_N\}$, we use a simple bag-of-words encoder:
    $f^{\text{enc}}(\{w_1,...,w_N\}) = \frac{1}{T}\sum_{i=1}^N \text{emb}(w_i)$.
We implement $\text{emb}(w_i)$ using an embedding layer followed by a linear transformation. The embedding layer is initialized with GloVe \citep{pennington2014glove} vectors and updated during training.
As the policy (decoder) we use a flat LSTM with 2 layers of 1024 LSTM units. The decoder hidden state is initialized with a linear transformation of $f^{\text{enc}}(\{w_1,...,w_T\})$.

\begin{table}[t]
  \footnotesize\centering
\begin{tabular}{@{}l|ccc|ccc@{}}
	\toprule
	 & \multicolumn{3}{c|}{\textbf{Validation}} & \multicolumn{3}{c}{\textbf{Test}} \\
	\textbf{Oracle} & \textbf{\textsc{Bleu}} &\textbf{F1} &\textbf{EM} & \textbf{\textsc{Bleu}} &\textbf{F1} &\textbf{EM}\\ 
	\midrule
	left-right & 46.6 & 0.910 & 0.230 & 46.3 & 0.903 & 0.208 \\
	uniform    & 44.7 & 0.968 & 0.209 & 44.3 & 0.960 & 0.197 \\
	annealed   & 46.8 & 0.960 & 0.230 & 46.0 & 0.950 & 0.212 \\
	\bottomrule
\end{tabular}%
\caption{\label{tbl:reorder_results}
Word Reordering results on Persona-Chat, reported according to \textsc{Bleu} score, F1 score, and percent exact match.}
\vspace{-1em}
\end{table}

\paragraph{Results.} \autoref{tbl:reorder_results} shows \textsc{Bleu}, F1 score, and exact match for policies trained with each oracle. The uniform and annealed policies outperform the left-right policy in F1 score (0.96 and 0.95 vs. 0.903). The policy trained using the annealed oracle also matches the left-right policy's performance in terms of \textsc{Bleu} score (46.0 vs. 46.3) and exact match (0.212 vs. 0.208). The model trained with the uniform policy does not fare as well on \textsc{Bleu} or exact match. See Appendix Figure 6 for example predictions.

\begin{figure}[t]
    \centering
    \includegraphics[width=0.8\columnwidth]{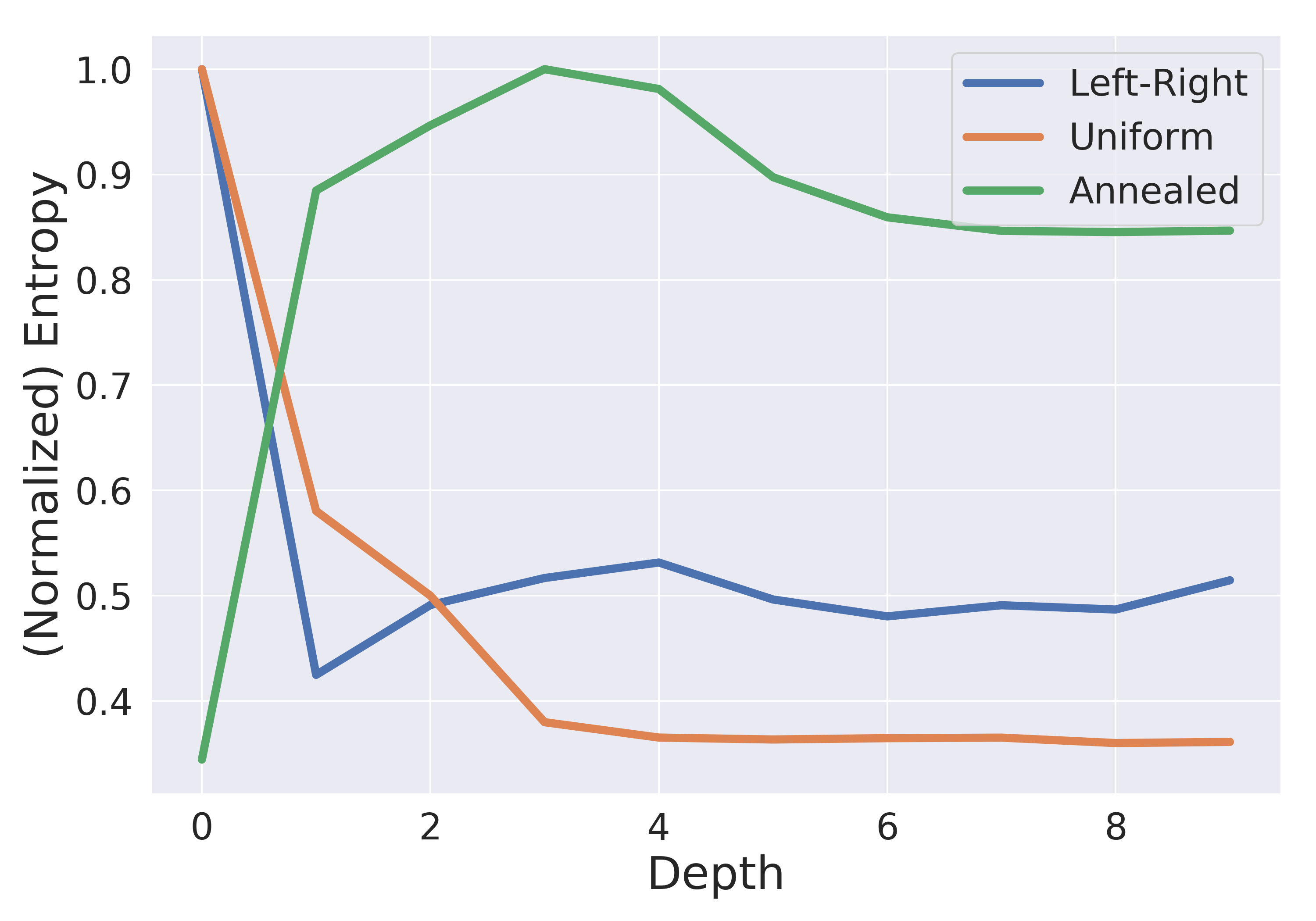}
    \vspace{-3mm}
    \caption{    \label{fig:entropy} Normalized entropy of $\pi(\cdot|s)$ as a function of tree depth for policies trained with each of the oracles. The anneal-trained policy, unlike the others, makes low entropy decisions early.}
    \vspace{-3mm}
\end{figure}

\paragraph{Easy-First Analysis.} 
\autoref{fig:entropy} shows the entropy of each model as a function of depth in the tree (normalized to fall in $[0,1]$). The left-right-trained policy has high entropy on the first word and then drops dramatically as additional conditioning from prior context kicks in. The uniform-trained policy exhibits similar behavior. The annealed-trained policy, however, makes its highest confidence (``easiest'') predictions at the beginning (consistent with \autoref{fig:unc_pos_heatmap}) and defers harder decisions until later.

\begin{table*}[t]
\centering\footnotesize
\begin{tabular}{@{}l|c@{ }cccc|c@{ }cccc@{}}
	\toprule
	& \multicolumn{5}{c|}{\textbf{Validation}}
	& \multicolumn{5}{c}{\textbf{Test}} \\
	\textbf{Oracle} 
	& \textbf{\textsc{Bleu}} & \textcolor{black!70}{\scriptsize (BP)} &\textbf{Meteor} &\textbf{YiSi}  &\textbf{Ribes} 
	& \textbf{\textsc{Bleu}} & \textcolor{black!70}{\scriptsize (BP)} &\textbf{Meteor} &\textbf{YiSi}  &\textbf{Ribes}\\ 
	\midrule
	left-right & 32.30 & \textcolor{black!70}{\scriptsize (0.95)} &             31.96 & 69.41 & 84.80 
	           & 28.00 & \textcolor{black!70}{\scriptsize (1.00)} & 30.10 & 65.22 & 82.29  \\[0.5em]
	uniform    & 24.50 & \textcolor{black!70}{\scriptsize (0.84)} &             27.98 & 66.40 & 82.66
	           & 21.40 & \textcolor{black!70}{\scriptsize (0.86)} & 26.40 & 62.41 & 80.00 \\[0.5em]
	annealed   & 26.80 & \textcolor{black!70}{\scriptsize (0.88)} &             29.67 & 67.88 & 83.61 
	           & 23.30 & \textcolor{black!70}{\scriptsize (0.91)} & 27.96 & 63.38 & 80.91 \\
	~+tree-encoding
	           & 28.00 & \textcolor{black!70}{\scriptsize (0.86)} & 30.15 & 68.43 & 84.36 
	           & 24.30 & \textcolor{black!70}{\scriptsize (0.91)} & 28.59 & 63.87 & 81.64 \\
	~+\angleend-tuning & 29.10 & \textcolor{black!70}{\scriptsize               (0.99)} & 31.00 & 68.81 & 83.51
               & 24.60 & \textcolor{black!70}{\scriptsize           (1.00)} & 29.30 & 64.18 & 80.53\\
	\bottomrule
\end{tabular}%
\caption{\label{tbl:nmt_results}
Results of machine translation experiments for different training oracles across four different evaluation metrics.}
    \vspace{-3mm}
\end{table*}

\subsection{Machine Translation} \label{sec:mt}

\paragraph{Dataset.} 
We evaluate the proposed models on IWSLT'16 German $\rightarrow{}$ English (196k pairs) translation task. The data sets consist of TED talks. We use TED tst2013 as a validation dataset and tst-2014 as test.

\paragraph{Model \& Training.} 
We use a Transformer policy, following the architecture of \cite{vaswani2017attention}. We use auxiliary \angleend prediction by introducing an additional output head, after observing a low brevity penalty in preliminary experiments. For the \angleend prediction threshold $\tau$ we use $0.5$, and also report a variant (+\angleend tuning) in which $\tau$ is tuned based on validation \textsc{Bleu} ($\tau=0.67)$. Finally, we report a variant which embeds each token by additionally encoding its path from the root (+tree-encoding) based on \citep{shiv2019novel}. See Appendix A.3 for additional details and results with a Bi-LSTM encoder-decoder architecture.

\paragraph{Results.} Results on validation and test data are in \autoref{tbl:nmt_results} according to four (very) different evaluation measures: \textsc{Bleu}, Meteor \citep{lavie2007meteor}, YiSi \citep{lo2018yisi}, and Ribes \citep{isozaki2010automatic}. First focusing on the non-monotonic models, we see that the annealed policy outperforms the uniform policy on all metrics, with tree-encoding yielding further gains. Adding \angleend tuning to the tree-encoding model decreases the Ribes score but improves the other metrics, notably increasing the \textsc{Bleu} brevity penalty.

Compared to the best non-monotonic model, the left-to-right model has superior performance according to \textsc{Bleu}. 
As previously observed \citep{callison2006re,wilks2008machine}, \textsc{Bleu} tends to strongly prefer models with left-to-right language models because it focuses on getting a large number of $4$-grams correct. 
The other three measures of translation quality are significantly less sensitive to exact word order and focus more on whether the ``semantics'' is preserved (for varying definitions of ``semantics''). For those, we see that the best annealed model is more competitive, typically within one percentage point of left-to-right. 


%% file: related.tex
Arguably one of the most successful approaches for generating discrete sequences, or sentences, is neural autoregressive modeling~\citep{sutskever2011generating,tomas2012statistical}. It has become {\it de facto} standard in machine translation~\citep{cho2014properties,sutskever2014sequence} and is widely studied for dialogue response generation~\citep{vinyals2015neural} as well as speech recognition~\citep{chorowski2015attention}. On the other hand, recent works have shown that it is possible to generate a sequence of discrete tokens in parallel by capturing strong dependencies among the tokens in a non-autoregressive way~\citep{gu2017non,lee2018deterministic,oord2017parallel}. \citet{stern2018blockwise} and \citet{wang2018semi} proposed to mix in these two paradigms and build a semi-autoregressive sequence generator, while largely sticking to left-to-right generation. Our proposal radically departs from these conventional approaches by building an algorithm that automatically captures a distinct generation order.

In (neural) language modeling, there is a long tradition of modeling the probability of a sequence as a tree or directed graph. For example, \citet{emami2005neural} proposed to factorize the probability over a sentence following its syntactic structure and train a neural network to model conditional distributions, which was followed more recently by \citet{zhang2015top} and by \citet{dyer2016recurrent}. This approach was applied to neural machine translation by \citet{eriguchi2017learning} and \citet{aharoni2017towards}. In all cases, these approaches require the availability of the ground-truth parse of a sentence or access to an external parser during training or inference time. This is unlike the proposed approach which does not require any such extra annotation or tool and learns to sequentially generate a sequence in an automatically determined non-monotonic order.


%% file: future.tex
We described an approach to generating text in non-monotonic orders that fall out naturally as the result of learning. We explored several different oracle models for imitation, and found that an annealed ``coaching'' oracle performed best, and learned a ``best-first'' strategy for language modeling, where it appears to significantly outperform alternatives. On a word re-ordering task, we found that this approach essentially ties left-to-right decoding, a rather promising finding given the decades of work on left-to-right models. 
In a machine translation setting, we found that the model learns to translate in a way that tends to preserve meaning but not n-grams.

There are several potentially interesting avenues for future work. One is to solve the ``learning to stop'' problem directly, rather than through an after-the-fact tuning step. Another is to better understand how to construct an oracle that generalizes well after mistakes have been made, in order to train off of the gold path(s).

Moreover, the proposed formulation of sequence generation by tree generation is limited to binary trees. 
It is possible to extend the proposed approach to $n$-ary trees by designing a policy to output up to $n+1$ decisions at each node, leading to up to $n$ child nodes. This would bring a set of generation orders, that could be captured by the proposed approach, which includes all projective dependency parses. A new oracle must be designed for $n$-ary trees, and we leave this as a follow-up work.

Finally, although the proposed approach indeed learns to sequentially generate a sequence in a non-monotonic order, it cannot consider all possible orders. It is due to the constraint that there cannot be any crossing of two edges when the nodes are arranged on a line following the inorder traversal, which we refer to as projective generation. Extending the proposed approach to non-projective generation, which we leave as future work, would expand the number of generation orders considered during learning.


%% file: appendix.tex
\appendix
\section{Additional Experiment Details and Results}
\subsection{Word Reordering} 
\paragraph{Model} The decoder is a 2-layer LSTM with 1024 hidden units, dropout of 0.0, based on a preliminary grid search of $n_{\text{layers}}\in\{1,2\},n_{\text{hidden}}\in\{512,1024,2048\},{\text{dropout}}\in\{0.0,0.2,0.5\}$. Word embeddings are initialized with GloVe vectors and updated during training. All presented Word Reordering results use greedy decoding.

\paragraph{Training} Each model was trained on a single GPU using a maximum of 500 epochs, batch size of 32, Adam \citet{} optimizer, gradient clipping with maximum $\ell_2$-norm of 1.0, and a learning rate starting at 0.001 and multiplied by a factor of 0.5 every 20 epochs. For evaluation we select the model state which had the highest validation BLEU score, which is evaluated after each training epoch.

\paragraph{Oracle} For $\pi_{\text{annealed}}^{*}$, $\beta$ is linearly annealed from 1.0 to 0.0 at a rate of 0.05 each epoch, after a burn-in period of 20 epochs in which $\beta$ is not decreased. We use greedy decoding when $\pi_{\text{coaching}}^*$ is selected at a roll-in step; we did not observe significant performance variations with stochastically sampling from $\pi_{\text{coaching}}^*$. These settings are based on a grid search of $\beta_{\text{rate}}\in\{0.01,0.05\},\beta_{\text{burn-in}}\in\{0,20\}, \text{coaching-rollin}\in\{\text{greedy},\text{stochastic}\}$ using the model selected in the \textbf{Model} section above.

\paragraph{Example Predictions}
Figure \ref{fig:bag_with_tree} shows example predictions from the validation set, including the generation order and underlying tree.

\subsection{Unconditional Generation} 
We use the same settings as the Word Reordering experiments, except we always use stochastic sampling from $\pi_{\text{coaching}}^*$ during roll-in. For evaluation we select the model state at the end of training. 

\paragraph{Unconditional Samples}
Samples in Tables \ref{tbl:unc-samples-short}-\ref{tbl:unc-samples-multi} are organized as `short' ($\leq$ 5th percentile), `average-length' (45-55th percentile), and `multi-sentence' ($\geq$ 3 punctuation tokens). 
Each image in Figures \ref{fig:unc_tree_anneal}, \ref{fig:unc_tree_unif}, and \ref{fig:unc_tree_lr} shows a sampled sentence, its underlying tree, and its generation order.

\begin{table}
\caption{Unconditional generation \textsc{BLEU} for various top-$k$ samplers and policies trained with the specified oracle.}
\label{tbl:unc_bleu}
\begin{center}
\begin{tabular}{@{}lc|ccc@{}}
	\toprule
	\textbf{Oracle} & \textbf{k} & \textbf{BLEU-2} & \textbf{BLEU-3} & \textbf{BLEU-4}\\ 
	\midrule
	$\pi_{\text{left-right}}^{*}$ & 10    &0.905 & 0.778 & 0.624\\
	                  & 100  &0.874 & 0.705 & 0.514\\
	                  & 1000  &0.853 & 0.665 & 0.466\\
	                  & all & 0.853 & 0.668 & 0.477 \\
	\midrule
	$\pi_{\text{uniform}}^{*}$ & 10  &0.966 & 0.906 & 0.788\\
	                  & 100 &0.916 & 0.751 & 0.544\\
	                  & 1000 &0.864 & 0.651 & 0.435\\
	                  & all & 0.831 & 0.609 & 0.395 \\
	\midrule
	$\pi_{\text{annealed}}^{*}$ & 10  &0.966 & 0.895 & 0.770\\
	                  & 100 &0.931 & 0.804 & 0.628\\
	                  & 1000 &0.907 & 0.765 & 0.585 \\
	                  & all & 0.894 & 0.740 & 0.549 \\
	\bottomrule
\end{tabular}%
\end{center}
\end{table}

\paragraph{Additional \textsc{BLEU} Scores}
Since absolute \textsc{BLEU} scores can vary by using a softmax temperature \citep{caccia2018language} or top-k sampler, we report additional scores for $k\in\{10, 100, 1000\}$ and \textsc{BLEU}-$\{2,3,4\}$ in Table \ref{tbl:unc_bleu}. 
Generally the policy trained with the annealed oracle achieves the highest metrics.

\begin{table*}[t]
\centering\footnotesize
\begin{tabular}{@{}l|c@{ }cccc|c@{ }cccc@{}}
	\toprule
	& \multicolumn{5}{c|}{\textbf{Validation}}
	& \multicolumn{5}{c}{\textbf{Test}} \\
	\textbf{Oracle} 
	& \textbf{\textsc{Bleu}} & \textcolor{black!70}{\scriptsize (BP)} &\textbf{Meteor} &\textbf{YiSi}  &\textbf{Ribes} 
	& \textbf{\textsc{Bleu}} & \textcolor{black!70}{\scriptsize (BP)} &\textbf{Meteor} &\textbf{YiSi}  &\textbf{Ribes}\\ 
	\midrule
	left-right & 29.47 & \textcolor{black!70}{\scriptsize (0.97)} & 29.66 & 52.03 & 82.55
	           & 26.23 & \textcolor{black!70}{\scriptsize (1.00)} & 27.87 & 47.58 & 79.85  \\[0.5em]
	uniform    & 14.97 & \textcolor{black!70}{\scriptsize (0.63)} & 21.76 & 41.62 & 77.70 
	           & 13.17 & \textcolor{black!70}{\scriptsize (0.64)} & 19.87 & 36.48 & 75.36 \\
	~+\angleend-tuning
	           & 18.79 
	           & \textcolor{black!70}{\scriptsize (0.89)} 
	           & 25.30 
	           & 46.23 
	           & 78.49 
	           & 17.68 
	           & \textcolor{black!70}{\scriptsize (0.96)} 
	           & 24.53 
	           & 42.46 
	           & 74.12
	           \\[0.5em]
	annealed   & 19.50 & \textcolor{black!70}{\scriptsize (0.71)} & 26.57 & 48.00 & 81.48 
	           & 16.94 & \textcolor{black!70}{\scriptsize (0.72)} & 23.15 & 42.39 & 78.99 \\
	~+\angleend-tuning
	           & 21.95 & \textcolor{black!70}{\scriptsize (0.90)} & 26.74 & 49.01 & 81.77 
	           & 19.19 & \textcolor{black!70}{\scriptsize (0.91)} & 25.24 & 43.98 & 79.24 \\
	\bottomrule
\end{tabular}%
\caption{\label{tbl:nmt_results}
LSTM Policy results for machine translation experiments.}
    \vspace{-3mm}
\end{table*}

\subsection{Machine Translation} 
\paragraph{Data and Preprocessing} We use the default Moses tokenizer script \citep{Koehn2007} and segment each word into a subword using BPE \citep{Sennrich2015} creating 40k tokens for both source and target. Similar to \citep{Bahdanau2015attention}, during training we filter sentence pairs that exceed 50 words.

\paragraph{Transformer Policy} The Transformer policy uses 4 layers, 4 attention heads, hidden dimension 256, feed-forward dimension 1024, and is trained with batch-size 32 and a learning rate $1\text{e}^{-5}$. For this model and experiment, we define an epoch as 1,000 model updates. The learning rate is divided by a factor of 1.1 every 100 epochs. For $\pi_{\text{annealed}}^*$, $\beta$ is linearly annealed from 1.0 to 0.0 at a rate of 0.01 each epoch, after a burn-in period of 100 epochs. We compute metrics after each validation epoch, and following training we select the model with the highest validation \textsc{Bleu}.

\paragraph{Loss with Auxiliary \angleend Predictor} A binary cross-entropy loss is used for the \angleend predictor for all time-steps, so that the total loss is $\mathcal{L}_{\text{bce}}(\pi^*, \pi_{\text{end}})+\mathcal{L}_{\text{KL}}(\pi^*, \pi)$ where $\mathcal{L}_{\text{KL}}$ is the loss from Section 3.2. For time-steps in which \angleend is sampled, $\mathcal{L}_{\text{KL}}$ is masked, since the policy's token distribution is not used when $a_t$ is \angleend. $\mathcal{L}_{\text{KL}}$ is averaged over time by summing the loss from unmasked time-steps, then dividing by the number of unmasked time-steps.

\paragraph{Tree Position Encodings} 
We use an additional \textit{tree position encoding}, based on \citep{shiv2019novel}, which may make it easier for the policy to identify and exploit structural relationships in the partially decoded tree. Each node is encoded using its path from the root, namely a sequence of left or right steps from parent to child. Each step is represented as a 2-dimensional binary vector ($[0,0]$ for the root, $[1,0]$ for left and $[0,1]$ for right), so that the path is a vector $e(a_i)\in \{0,1\}^{2*\text{max-depth}}$ after zero-padding. 
Finally, $e(a_i)$ is multiplied element-wise by a geometric series of a learned parameter $p$, that is, $e(a_i) \cdot \left[1, p, p, p^2,p^3,...\right]$. We only use this approach with the Transformer policy. 

\paragraph{Additional LSTM Policy}  Results are shown in \autoref{tbl:nmt_results}. We use a bi-directional LSTM encoder-decoder architecture that has a single layer of size 512, with global concat attention \citep{Thang2015attention}. The learning rate is initialized to 0.001 and multiplied by a factor of 0.5 on a fixed interval. 

\begin{table*}
\caption{\textbf{Short} (left) and \textbf{Average-Length} (right) unconditional samples from policies trained on Persona-Chat.}
\label{tbl:unc-samples-short}
\begin{tabular*}{\textwidth}{ll@{\hskip 2cm}l}
	\toprule
left-right & i can drive you alone . & do you like to test your voice to a choir ?\\
& yeah it is very important . & no pets , on the subject in my family , yes .\\
& i am a am nurse . & cool . i have is also a cat named cow .\\
& do you actually enjoy it ? & i am doing good taking a break from working on it .\\
& what pair were you in ? & i do not have one , do you have any pets ?\\

\midrule
uniform & good just normal people around .& just that is for a while . and yourself right now ?\\
& you run the hills right ? & i am freelance a writer but i am a writer .\\
& i am great yourself ?& that is so sad . do you have a free time ?\\
& i work 12 hours .& yes i do not like pizza which is amazing lol .\\
& do you go to hockey ?& since the gym did not bother me many years ago .\\

\midrule
annealed & are you ? i am .& yeah it can be . what is your favorite color ?\\
& i like to be talented .& i do not have dogs . they love me here .\\
& how are you doing buddy ?& no kids . . . i am . . you ?\\
& i like healthy foods .& that is interesting . i am just practicing my piano degree .\\
& i love to eat .& yea it is , you need to become a real nerd !\\
\midrule
\end{tabular*}
\end{table*}

\begin{table*}
\caption{\textbf{Multi-sentence} unconditional samples from policies trained on Persona-Chat.}
\label{tbl:unc-samples-multi}
\begin{tabular*}{\textwidth}{ll@{\extracolsep{\fill}}}
	\toprule
left-right & nice ! i think i will get a jump blade again . have you done that at it ?\\
& great . what kinds of food do you like best ? i love italian food .\\
& wow . bike ride is my thing . i do nothing for kids .\\
& i am alright . my mom makes work and work as a nurse . that is what i do for work .\\
& that is awesome . i need to lose weight . i want to start a food place someday . \\

\midrule
uniform & love meat . or junk food . i sometimes go too much i make . avoid me unhealthy .\\
& does not kill anyone that can work around a lot of animals ? you ? i like trains .\\
& baby ? it will it all here . that is the workforce .\\
& i am good , thank you . i love my sci fi stories . i write books .\\
& i am well . thank you . my little jasper is new .\\

\midrule
annealed & i am definitely a kid . are you ? i am 10 !\\
& i am in michigan state . . that is a grand state .\\
& that is good . i work as a pharmacist in florida . . .\\
& how are you ? wanna live in san fran ! i love it .\\
& well that is awesome ! i do crosswords ! that is cool .\\
\midrule
\end{tabular*}
\end{table*}
%
\newcounter{myrow}
\newcounter{mycolumn}

\begin{figure*}
\caption{Unconditional samples from a policy trained with $\pi_{\text{annealed}}^{*}$.}
\label{fig:unc_tree_anneal}
\vspace{0.2cm}
\begin{center}
\whiledo{\themyrow<4}{
     \whiledo{\themycolumn<2}{
     \noindent\includegraphics[width=0.48\textwidth]{images/unc/tree_anneal_coaching_\themyrow_\themycolumn_k-1.png}
     \vspace{-1cm}
     \stepcounter{mycolumn}
   }\setcounter{mycolumn}{0}\\\stepcounter{myrow}
}
\end{center}
\end{figure*}
\setcounter{mycolumn}{0}
\setcounter{myrow}{0}
\begin{figure*}
\caption{Unconditional samples from a policy trained with $\pi_{\text{uniform}}^{*}$.}
\label{fig:unc_tree_unif}
\vspace{0.2cm}
\begin{center}
\whiledo{\themyrow<4}{
     \whiledo{\themycolumn<2}{
     \noindent\includegraphics[width=0.48\textwidth]{images/unc/tree_uniform_\themyrow_\themycolumn_k-1.png}
     \vspace{-1cm}
     \stepcounter{mycolumn}
   }\setcounter{mycolumn}{0}\\\stepcounter{myrow}
}
\end{center}
\end{figure*}
\setcounter{mycolumn}{0}
\setcounter{myrow}{0}
\begin{figure*}
\caption{Unconditional samples from a policy trained with $\pi_{\text{left-right}}^{*}$.}
\label{fig:unc_tree_lr}
\vspace{0.2cm}
\begin{center}
\whiledo{\themyrow<4}{
     \whiledo{\themycolumn<2}{
     \noindent\includegraphics[width=0.48\textwidth]{images/unc/tree_leftright_\themyrow_\themycolumn_k-1.png}
     \vspace{-0.5cm}
     \stepcounter{mycolumn}
   }\setcounter{mycolumn}{0}\\\stepcounter{myrow}
}
\end{center}
\end{figure*}

\setcounter{mycolumn}{0}
\setcounter{myrow}{0}
\begin{figure*}
\caption{Word Reordering Examples. The columns show policies trained with $\pi_{\text{left-right}}^{*}$, $\pi_{\text{uniform}}^{*}$, and $\pi_{\text{annealed}}^{*}$, respectively.}
\label{fig:bag_with_tree}
\vspace{0.2cm}
\begin{center}
\whiledo{\themyrow<4}{
     \whiledo{\themycolumn<3}{
     \ifthenelse{\themycolumn=0}{\noindent\includegraphics[width=0.31\textwidth]{images/bag/bag_leftright_\themyrow_\themycolumn.png}}{}
     \ifthenelse{\themycolumn=1}{\noindent\includegraphics[width=0.31\textwidth]{images/bag/bag_uniform_\themyrow_\themycolumn.png}}{}
     \ifthenelse{\themycolumn=2}{\noindent\includegraphics[width=0.31\textwidth]{images/bag/bag_anneal_coaching_\themyrow_\themycolumn.png}}{}
     \stepcounter{mycolumn}
   }\setcounter{mycolumn}{0}\\\stepcounter{myrow}
}
\end{center}
\end{figure*}

\setcounter{mycolumn}{0}
\setcounter{myrow}{1}
\begin{figure*}
\caption{Translation outputs from a policy trained with $\pi_{\text{annealed}}^{*}$ on the test set.}
\vspace{0.2cm}
\begin{center}
\whiledo{\themyrow<5}{
     \whiledo{\themycolumn<2}{
     \ifthenelse{\themycolumn=0}{\noindent\includegraphics[width=0.31\textwidth]{images/translation/anneal/translation_\themyrow_\themycolumn.png}}{}
     \ifthenelse{\themycolumn=1}{\noindent\includegraphics[width=0.31\textwidth]{images/translation/anneal/translation_\themyrow_\themycolumn.png}}{}
     \stepcounter{mycolumn}
   }
   \setcounter{mycolumn}{0}\\
   \stepcounter{myrow}
}
\end{center}
\end{figure*}

\setcounter{mycolumn}{0}
\setcounter{myrow}{1}
\begin{figure*}
\caption{Translation outputs from a policy trained with $\pi_{\text{uniform}}^{*}$ on the test set.}
\vspace{0.2cm}
\begin{center}
\whiledo{\themyrow<5}{
     \whiledo{\themycolumn<2}{
     \ifthenelse{\themycolumn=0}{\noindent\includegraphics[width=0.31\textwidth]{images/translation/uniform/translation_\themyrow_\themycolumn.png}}{}
     \ifthenelse{\themycolumn=1}{\noindent\includegraphics[width=0.31\textwidth]{images/translation/uniform/translation_\themyrow_\themycolumn.png}}{}
     \stepcounter{mycolumn}
   }
   \setcounter{mycolumn}{0}\\
   \stepcounter{myrow}
}
\end{center}
\end{figure*}

\setcounter{mycolumn}{0}
\setcounter{myrow}{1}
\begin{figure*}
\caption{Translation outputs from a policy trained with $\pi_{\text{leftright}}^{*}$ on the test set.}
\vspace{0.2cm}
\begin{center}
\whiledo{\themyrow<5}{
     \whiledo{\themycolumn<2}{
     \ifthenelse{\themycolumn=0}{\noindent\includegraphics[width=0.31\textwidth]{images/translation/leftright/translation_\themyrow_\themycolumn.png}}{}
     \ifthenelse{\themycolumn=1}{\noindent\includegraphics[width=0.31\textwidth]{images/translation/leftright/translation_\themyrow_\themycolumn.png}}{}
     \stepcounter{mycolumn}
   }
   \setcounter{mycolumn}{0}\\
   \stepcounter{myrow}
}
\end{center}
\end{figure*}

%% file: nonmonotone.bbl
\begin{thebibliography}{66}
\providecommand{\natexlab}[1]{#1}
\providecommand{\url}[1]{\texttt{#1}}
\expandafter\ifx\csname urlstyle\endcsname\relax
  \providecommand{\doi}[1]{doi: #1}\else
  \providecommand{\doi}{doi: \begingroup \urlstyle{rm}\Url}\fi

\bibitem[Aharoni \& Goldberg(2017)Aharoni and Goldberg]{aharoni2017towards}
Aharoni, R. and Goldberg, Y.
\newblock Towards string-to-tree neural machine translation.
\newblock \emph{arXiv preprint arXiv:1704.04743}, 2017.

\bibitem[Alvarez-Melis \& Jaakkola(2017)Alvarez-Melis and
  Jaakkola]{alvarez2016tree}
Alvarez-Melis, D. and Jaakkola, T.~S.
\newblock Tree-structured decoding with doubly-recurrent neural networks.
\newblock \emph{International Conference on Learning Representations (ICLR)},
  2017.

\bibitem[Bahdanau et~al.(2015{\natexlab{a}})Bahdanau, Cho, and
  Bengio]{Bahdanau2015attention}
Bahdanau, D., Cho, K., and Bengio, Y.
\newblock Neural machine translation by jointly learning to align and
  translate.
\newblock \emph{In International Conference on Learning Representations},
  2015{\natexlab{a}}.

\bibitem[Bahdanau et~al.(2015{\natexlab{b}})Bahdanau, Serdyuk, Brakel, Ke,
  Chorowski, Courville, and Bengio]{bahdanau2015task}
Bahdanau, D., Serdyuk, D., Brakel, P., Ke, N.~R., Chorowski, J., Courville, A.,
  and Bengio, Y.
\newblock Task loss estimation for sequence prediction.
\newblock \emph{arXiv preprint arXiv:1511.06456}, 2015{\natexlab{b}}.

\bibitem[Bahdanau et~al.(2016)Bahdanau, Brakel, Xu, Goyal, Lowe, Pineau,
  Courville, and Bengio]{bahdanau2016actor}
Bahdanau, D., Brakel, P., Xu, K., Goyal, A., Lowe, R., Pineau, J., Courville,
  A., and Bengio, Y.
\newblock An actor-critic algorithm for sequence prediction.
\newblock \emph{arXiv preprint arXiv:1607.07086}, 2016.

\bibitem[Bahl et~al.(1983)Bahl, Jelinek, and Mercer]{bahl1983maximum}
Bahl, L.~R., Jelinek, F., and Mercer, R.~L.
\newblock A maximum likelihood approach to continuous speech recognition.
\newblock \emph{IEEE transactions on pattern analysis and machine
  intelligence}, 5\penalty0 (2):\penalty0 179--190, 1983.

\bibitem[Battaglia et~al.(2018)Battaglia, Hamrick, Bapst, Sanchez-Gonzalez,
  Zambaldi, Malinowski, Tacchetti, Raposo, Santoro, Faulkner,
  et~al.]{battaglia2018relational}
Battaglia, P.~W., Hamrick, J.~B., Bapst, V., Sanchez-Gonzalez, A., Zambaldi,
  V., Malinowski, M., Tacchetti, A., Raposo, D., Santoro, A., Faulkner, R.,
  et~al.
\newblock Relational inductive biases, deep learning, and graph networks.
\newblock \emph{arXiv preprint arXiv:1806.01261}, 2018.

\bibitem[Bengio et~al.(2003)Bengio, Ducharme, Vincent, and
  Jauvin]{bengio2003neural}
Bengio, Y., Ducharme, R., Vincent, P., and Jauvin, C.
\newblock A neural probabilistic language model.
\newblock \emph{Journal of machine learning research}, 3\penalty0
  (Feb):\penalty0 1137--1155, 2003.

\bibitem[Berglund et~al.(2015)Berglund, Raiko, Honkala, K{\"a}rkk{\"a}inen,
  Vetek, and Karhunen]{berglund2015bidirectional}
Berglund, M., Raiko, T., Honkala, M., K{\"a}rkk{\"a}inen, L., Vetek, A., and
  Karhunen, J.~T.
\newblock Bidirectional recurrent neural networks as generative models.
\newblock In \emph{Advances in Neural Information Processing Systems}, pp.\
  856--864, 2015.

\bibitem[Bowman et~al.(2016)Bowman, Gauthier, Rastogi, Gupta, Manning, and
  Potts]{bowman2016fast}
Bowman, S.~R., Gauthier, J., Rastogi, A., Gupta, R., Manning, C.~D., and Potts,
  C.
\newblock A fast unified model for parsing and sentence understanding.
\newblock \emph{arXiv preprint arXiv:1603.06021}, 2016.

\bibitem[Bronstein et~al.(2017)Bronstein, Bruna, LeCun, Szlam, and
  Vandergheynst]{bronstein2017geometric}
Bronstein, M.~M., Bruna, J., LeCun, Y., Szlam, A., and Vandergheynst, P.
\newblock Geometric deep learning: going beyond euclidean data.
\newblock \emph{IEEE Signal Processing Magazine}, 34\penalty0 (4):\penalty0
  18--42, 2017.

\bibitem[Brown et~al.(1990)Brown, Cocke, Pietra, Pietra, Jelinek, Lafferty,
  Mercer, and Roossin]{brown1990statistical}
Brown, P.~F., Cocke, J., Pietra, S. A.~D., Pietra, V. J.~D., Jelinek, F.,
  Lafferty, J.~D., Mercer, R.~L., and Roossin, P.~S.
\newblock A statistical approach to machine translation.
\newblock \emph{Comput. Linguist.}, 16\penalty0 (2):\penalty0 79--85, June
  1990.
\newblock ISSN 0891-2017.
\newblock URL \url{http://dl.acm.org/citation.cfm?id=92858.92860}.

\bibitem[Caccia et~al.(2018)Caccia, Caccia, Fedus, Larochelle, Pineau, {Charlin
  MILA}, and Montr{\'{e}}al]{caccia2018language}
Caccia, M., Caccia, L., Fedus, W., Larochelle, H., Pineau, J., {Charlin MILA},
  L., and Montr{\'{e}}al, H.
\newblock Language gans falling short.
\newblock \emph{arXiv preprint 1811.02549}, 2018.
\newblock URL \url{https://arxiv.org/pdf/1811.02549.pdf}.

\bibitem[Callison-Burch et~al.(2006)Callison-Burch, Osborne, and
  Koehn]{callison2006re}
Callison-Burch, C., Osborne, M., and Koehn, P.
\newblock Re-evaluation the role of bleu in machine translation research.
\newblock In \emph{11th Conference of the European Chapter of the Association
  for Computational Linguistics}, 2006.

\bibitem[Chang et~al.(2015)Chang, Krishnamurthy, Agarwal, Daum{\'e}~III, and
  Langford]{chang2015learning}
Chang, K.-W., Krishnamurthy, A., Agarwal, A., Daum{\'e}~III, H., and Langford,
  J.
\newblock Learning to search better than your teacher.
\newblock \emph{arXiv preprint arXiv:1502.02206}, 2015.

\bibitem[Cheng et~al.(2018)Cheng, Yan, Wagener, and Boots]{cheng2018fast}
Cheng, C.-A., Yan, X., Wagener, N., and Boots, B.
\newblock {Fast Policy Learning through Imitation and Reinforcement}.
\newblock \emph{arXiv preprint 1805.10413}, 2018.
\newblock URL \url{https://arxiv.org/pdf/1805.10413.pdf}.

\bibitem[Chiang(2012)]{chiang2012hope}
Chiang, D.
\newblock Hope and fear for discriminative training of statistical translation
  models.
\newblock \emph{Journal of Machine Learning Research}, 13\penalty0
  (Apr):\penalty0 1159--1187, 2012.

\bibitem[Cho et~al.(2014{\natexlab{a}})Cho, Van~Merri{\"e}nboer, Bahdanau, and
  Bengio]{cho2014properties}
Cho, K., Van~Merri{\"e}nboer, B., Bahdanau, D., and Bengio, Y.
\newblock On the properties of neural machine translation: Encoder-decoder
  approaches.
\newblock \emph{arXiv preprint arXiv:1409.1259}, 2014{\natexlab{a}}.

\bibitem[Cho et~al.(2014{\natexlab{b}})Cho, Van~Merri{\"e}nboer, Gulcehre,
  Bahdanau, Bougares, Schwenk, and Bengio]{cho2014learning}
Cho, K., Van~Merri{\"e}nboer, B., Gulcehre, C., Bahdanau, D., Bougares, F.,
  Schwenk, H., and Bengio, Y.
\newblock Learning phrase representations using rnn encoder-decoder for
  statistical machine translation.
\newblock \emph{arXiv preprint arXiv:1406.1078}, 2014{\natexlab{b}}.

\bibitem[Cho et~al.(2015)Cho, Courville, and Bengio]{cho2015describing}
Cho, K., Courville, A., and Bengio, Y.
\newblock Describing multimedia content using attention-based encoder-decoder
  networks.
\newblock \emph{IEEE Transactions on Multimedia}, 17\penalty0 (11):\penalty0
  1875--1886, 2015.

\bibitem[Chorowski et~al.(2015)Chorowski, Bahdanau, Serdyuk, Cho, and
  Bengio]{chorowski2015attention}
Chorowski, J.~K., Bahdanau, D., Serdyuk, D., Cho, K., and Bengio, Y.
\newblock Attention-based models for speech recognition.
\newblock In \emph{Advances in neural information processing systems}, pp.\
  577--585, 2015.

\bibitem[Cleeremans et~al.(1989)Cleeremans, Servan-Schreiber, and
  McClelland]{cleeremans1989finite}
Cleeremans, A., Servan-Schreiber, D., and McClelland, J.~L.
\newblock Finite state automata and simple recurrent networks.
\newblock \emph{Neural computation}, 1\penalty0 (3):\penalty0 372--381, 1989.

\bibitem[Daum{\'e} et~al.(2009)Daum{\'e}, Langford, and Marcu]{daume2009search}
Daum{\'e}, H., Langford, J., and Marcu, D.
\newblock Search-based structured prediction.
\newblock \emph{Machine learning}, 75\penalty0 (3):\penalty0 297--325, 2009.

\bibitem[Daum\'e(2009)]{daume09unsearn}
Daum\'e, III, H.
\newblock Unsupervised search-based structured prediction.
\newblock In \emph{International Conference on Machine Learning (ICML)},
  Montreal, Canada, 2009.

\bibitem[Dyer et~al.(2015)Dyer, Ballesteros, Ling, Matthews, and
  Smith]{dyer2015transition}
Dyer, C., Ballesteros, M., Ling, W., Matthews, A., and Smith, N.~A.
\newblock Transition-based dependency parsing with stack long short-term
  memory.
\newblock \emph{arXiv preprint arXiv:1505.08075}, 2015.

\bibitem[Dyer et~al.(2016)Dyer, Kuncoro, Ballesteros, and
  Smith]{dyer2016recurrent}
Dyer, C., Kuncoro, A., Ballesteros, M., and Smith, N.~A.
\newblock Recurrent neural network grammars.
\newblock \emph{arXiv preprint arXiv:1602.07776}, 2016.

\bibitem[Emami \& Jelinek(2005)Emami and Jelinek]{emami2005neural}
Emami, A. and Jelinek, F.
\newblock A neural syntactic language model.
\newblock \emph{Machine learning}, 60\penalty0 (1-3):\penalty0 195--227, 2005.

\bibitem[Eriguchi et~al.(2017)Eriguchi, Tsuruoka, and
  Cho]{eriguchi2017learning}
Eriguchi, A., Tsuruoka, Y., and Cho, K.
\newblock Learning to parse and translate improves neural machine translation.
\newblock \emph{arXiv preprint arXiv:1702.03525}, 2017.

\bibitem[Forcada \& {\~{N}eco}(1997)Forcada and
  {\~{N}eco}]{forcada1997recursive}
Forcada, M.~L. and {\~{N}eco}, R.
\newblock Recursive hetero-associative memories for translation.
\newblock In \emph{International Work-Conference on Artificial Neural
  Networks}, 1997.

\bibitem[Ford et~al.(2018)Ford, Duckworth, Norouzi, and
  Dahl]{ford2018importance}
Ford, N., Duckworth, D., Norouzi, M., and Dahl, G.~E.
\newblock The importance of generation order in language modeling.
\newblock \emph{arXiv preprint arXiv:1808.07910}, 2018.

\bibitem[Goldberg \& Elhadad(2010)Goldberg and Elhadad]{goldberg2010efficient}
Goldberg, Y. and Elhadad, M.
\newblock An efficient algorithm for easy-first non-directional dependency
  parsing.
\newblock In \emph{Human Language Technologies: The 2010 Annual Conference of
  the North American Chapter of the Association for Computational Linguistics},
  pp.\  742--750. Association for Computational Linguistics, 2010.

\bibitem[Gu et~al.(2017)Gu, Bradbury, Xiong, Li, and Socher]{gu2017non}
Gu, J., Bradbury, J., Xiong, C., Li, V.~O., and Socher, R.
\newblock Non-autoregressive neural machine translation.
\newblock \emph{arXiv preprint arXiv:1711.02281}, 2017.

\bibitem[Hazan et~al.(2010)Hazan, Keshet, and McAllester]{hazan2010direct}
Hazan, T., Keshet, J., and McAllester, D.~A.
\newblock Direct loss minimization for structured prediction.
\newblock In \emph{Advances in Neural Information Processing Systems}, pp.\
  1594--1602, 2010.

\bibitem[He et~al.(2012)He, Eisner, and Daume]{he2012imitation}
He, H., Eisner, J., and Daume, H.
\newblock Imitation learning by coaching.
\newblock In \emph{Advances in Neural Information Processing Systems}, pp.\
  3149--3157, 2012.

\bibitem[Hochreiter \& Schmidhuber(1997)Hochreiter and
  Schmidhuber]{hochreiter1997long}
Hochreiter, S. and Schmidhuber, J.
\newblock Long short-term memory.
\newblock \emph{Neural computation}, 9\penalty0 (8):\penalty0 1735--1780, 1997.

\bibitem[Isozaki et~al.(2010)Isozaki, Hirao, Duh, Sudoh, and
  Tsukada]{isozaki2010automatic}
Isozaki, H., Hirao, T., Duh, K., Sudoh, K., and Tsukada, H.
\newblock Automatic evaluation of translation quality for distant language
  pairs.
\newblock In \emph{Proceedings of the 2010 Conference on Empirical Methods in
  Natural Language Processing}, pp.\  944--952. Association for Computational
  Linguistics, 2010.

\bibitem[Koehn et~al.(2007)Koehn, Hoang, Birch, Callison-Burch, Federico,
  Bertoldi, Cowan, Shen, Moran, Zens, Dyer, Bojar, Constantin, and
  Herbst]{Koehn2007}
Koehn, P., Hoang, H., Birch, A., Callison-Burch, C., Federico, M., Bertoldi,
  N., Cowan, B., Shen, W., Moran, C., Zens, R., Dyer, C., Bojar, O.,
  Constantin, A., and Herbst, E.
\newblock Moses: Open source toolkit for statistical machine translation.
\newblock In \emph{Proceedings of the 45th Annual Meeting of the ACL on
  Interactive Poster and Demonstration Sessions}, pp.\  177--180. Association
  for Computational Linguistics, 2007.

\bibitem[Lavie \& Agarwal(2007)Lavie and Agarwal]{lavie2007meteor}
Lavie, A. and Agarwal, A.
\newblock Meteor: An automatic metric for mt evaluation with high levels of
  correlation with human judgments.
\newblock In \emph{Proceedings of the Second Workshop on Statistical Machine
  Translation}, pp.\  228--231. Association for Computational Linguistics,
  2007.

\bibitem[Leblond et~al.(2018)Leblond, Alayrac, Osokin, and
  Lacoste-Julien]{leblond2018searnn}
Leblond, R., Alayrac, J.-B., Osokin, A., and Lacoste-Julien, S.
\newblock {SeaRNN}: Training {RNN}s with global-local losses.
\newblock In \emph{ICLR}, 2018.

\bibitem[Lee et~al.(2018)Lee, Mansimov, and Cho]{lee2018deterministic}
Lee, J., Mansimov, E., and Cho, K.
\newblock Deterministic non-autoregressive neural sequence modeling by
  iterative refinement.
\newblock \emph{arXiv preprint arXiv:1802.06901}, 2018.

\bibitem[Lo(2018)]{lo2018yisi}
Lo, C.
\newblock {YiSi}: A semantic machine translation evaluation metric for
  evaluating languages with different levels of available resources.
\newblock Unpublished, 2018.
\newblock URL \url{http://chikiu-jackie-lo.org/home/index.php/yisi}.

\bibitem[Luong et~al.(2015)Luong, Pham, and Manning]{Thang2015attention}
Luong, T., Pham, H., and Manning, C.~D.
\newblock Effective approaches to attention-based neural machine translation.
\newblock In \emph{Proceedings of the 2015 Conference on Empirical Methods in
  Natural Language Processing}, pp.\  1412--1421. Association for Computational
  Linguistics, 2015.

\bibitem[Oord et~al.(2017)Oord, Li, Babuschkin, Simonyan, Vinyals, Kavukcuoglu,
  Driessche, Lockhart, Cobo, Stimberg, et~al.]{oord2017parallel}
Oord, A. v.~d., Li, Y., Babuschkin, I., Simonyan, K., Vinyals, O., Kavukcuoglu,
  K., Driessche, G. v.~d., Lockhart, E., Cobo, L.~C., Stimberg, F., et~al.
\newblock Parallel wavenet: Fast high-fidelity speech synthesis.
\newblock \emph{arXiv preprint arXiv:1711.10433}, 2017.

\bibitem[Pennington et~al.(2014)Pennington, Socher, and
  Manning]{pennington2014glove}
Pennington, J., Socher, R., and Manning, C.~D.
\newblock Glove: Global vectors for word representation.
\newblock In \emph{Empirical Methods in Natural Language Processing (EMNLP)},
  pp.\  1532--1543, 2014.
\newblock URL \url{http://www.aclweb.org/anthology/D14-1162}.

\bibitem[Ranzato et~al.(2015)Ranzato, Chopra, Auli, and
  Zaremba]{ranzato2015sequence}
Ranzato, M., Chopra, S., Auli, M., and Zaremba, W.
\newblock Sequence level training with recurrent neural networks.
\newblock \emph{arXiv preprint arXiv:1511.06732}, 2015.

\bibitem[Ross \& Bagnell(2014)Ross and Bagnell]{ross2014reinforcement}
Ross, S. and Bagnell, J.~A.
\newblock Reinforcement and imitation learning via interactive no-regret
  learning.
\newblock \emph{arXiv preprint arXiv:1406.5979}, 2014.

\bibitem[Ross et~al.(2011)Ross, Gordon, and Bagnell]{ross2011reduction}
Ross, S., Gordon, G., and Bagnell, D.
\newblock A reduction of imitation learning and structured prediction to
  no-regret online learning.
\newblock In \emph{Proceedings of the fourteenth international conference on
  artificial intelligence and statistics}, pp.\  627--635, 2011.

\bibitem[Schmaltz et~al.(2016)Schmaltz, Rush, and Shieber]{schmaltz2016word}
Schmaltz, A., Rush, A.~M., and Shieber, S.
\newblock Word ordering without syntax.
\newblock In \emph{Proceedings of the 2016 Conference on Empirical Methods in
  Natural Language Processing}, pp.\  2319--2324. Association for Computational
  Linguistics, 2016.
\newblock \doi{10.18653/v1/D16-1255}.
\newblock URL \url{http://aclweb.org/anthology/D16-1255}.

\bibitem[Sennrich et~al.(2015)Sennrich, Haddow, and Birch]{Sennrich2015}
Sennrich, R., Haddow, B., and Birch, A.
\newblock Neural machine translation of rare words with subword units.
\newblock \emph{arXiv preprint arXiv:1508.07909}, 2015.

\bibitem[Shiv \& Quirk(2019)Shiv and Quirk]{shiv2019novel}
Shiv, V.~L. and Quirk, C.
\newblock Novel positional encodings to enable tree-structured transformers.
\newblock 2019.
\newblock URL \url{https://openreview.net/forum?id=SJerEhR5Km}.

\bibitem[Stern et~al.(2018)Stern, Shazeer, and Uszkoreit]{stern2018blockwise}
Stern, M., Shazeer, N., and Uszkoreit, J.
\newblock Blockwise parallel decoding for deep autoregressive models.
\newblock In \emph{Advances in Neural Information Processing Systems}, pp.\
  10107--10116, 2018.

\bibitem[Stoyanov \& Eisner(2012)Stoyanov and Eisner]{stoyanov2012easy}
Stoyanov, V. and Eisner, J.
\newblock Easy-first coreference resolution.
\newblock \emph{Proceedings of COLING 2012}, pp.\  2519--2534, 2012.

\bibitem[Sutskever et~al.(2011)Sutskever, Martens, and
  Hinton]{sutskever2011generating}
Sutskever, I., Martens, J., and Hinton, G.~E.
\newblock Generating text with recurrent neural networks.
\newblock In \emph{Proceedings of the 28th International Conference on Machine
  Learning (ICML-11)}, pp.\  1017--1024, 2011.

\bibitem[Sutskever et~al.(2014)Sutskever, Vinyals, and
  Le]{sutskever2014sequence}
Sutskever, I., Vinyals, O., and Le, Q.~V.
\newblock Sequence to sequence learning with neural networks.
\newblock In \emph{Advances in neural information processing systems}, pp.\
  3104--3112, 2014.

\bibitem[Tai et~al.(2015)Tai, Socher, and Manning]{tai2015improved}
Tai, K.~S., Socher, R., and Manning, C.~D.
\newblock Improved semantic representations from tree-structured long
  short-term memory networks.
\newblock \emph{arXiv preprint arXiv:1503.00075}, 2015.

\bibitem[Tomas(2012)]{tomas2012statistical}
Tomas, M.
\newblock Statistical language models based on neural networks.
\newblock \emph{Brno University of Technology}, 2012.

\bibitem[Tsuruoka \& Tsujii(2005)Tsuruoka and
  Tsujii]{tsuruoka2005bidirectional}
Tsuruoka, Y. and Tsujii, J.
\newblock Bidirectional inference with the easiest-first strategy for tagging
  sequence data.
\newblock In \emph{Proceedings of the conference on human language technology
  and empirical methods in natural language processing}, pp.\  467--474.
  Association for Computational Linguistics, 2005.

\bibitem[Vaswani et~al.(2017)Vaswani, Shazeer, Parmar, Uszkoreit, Jones, Gomez,
  Kaiser, and Polosukhin]{vaswani2017attention}
Vaswani, A., Shazeer, N., Parmar, N., Uszkoreit, J., Jones, L., Gomez, A.~N.,
  Kaiser, L.~u., and Polosukhin, I.
\newblock Attention is all you need.
\newblock In Guyon, I., Luxburg, U.~V., Bengio, S., Wallach, H., Fergus, R.,
  Vishwanathan, S., and Garnett, R. (eds.), \emph{Advances in Neural
  Information Processing Systems 30}, pp.\  5998--6008. Curran Associates,
  Inc., 2017.
\newblock URL
  \url{http://papers.nips.cc/paper/7181-attention-is-all-you-need.pdf}.

\bibitem[Vinyals \& Le(2015)Vinyals and Le]{vinyals2015neural}
Vinyals, O. and Le, Q.
\newblock A neural conversational model.
\newblock \emph{arXiv preprint arXiv:1506.05869}, 2015.

\bibitem[Wang et~al.(2018)Wang, Zhang, and Chen]{wang2018semi}
Wang, C., Zhang, J., and Chen, H.
\newblock Semi-autoregressive neural machine translation.
\newblock \emph{arXiv preprint arXiv:1808.08583}, 2018.

\bibitem[Welleck et~al.(2018)Welleck, Yao, Gai, Mao, Zhang, and
  Cho]{welleck2018loss}
Welleck, S., Yao, Z., Gai, Y., Mao, J., Zhang, Z., and Cho, K.
\newblock Loss functions for multiset prediction.
\newblock In \emph{Advances in Neural Information Processing Systems}, pp.\
  5788--5797, 2018.

\bibitem[Wilks(2008)]{wilks2008machine}
Wilks, Y.
\newblock \emph{Machine translation: its scope and limits}.
\newblock Springer Science \& Business Media, 2008.

\bibitem[Yu et~al.(2016)Yu, Zhang, Wang, and Yu]{yu2016seqgan}
Yu, L., Zhang, W., Wang, J., and Yu, Y.
\newblock Seqgan: Sequence generative adversarial nets with policy gradient.
\newblock \emph{CoRR}, abs/1609.05473, 2016.
\newblock URL
  \url{http://dblp.uni-trier.de/db/journals/corr/corr1609.html#YuZWY16}.

\bibitem[Zhang et~al.(2018)Zhang, Dinan, Urbanek, Szlam, Kiela, and
  Weston]{zhang2018personalizing}
Zhang, S., Dinan, E., Urbanek, J., Szlam, A., Kiela, D., and Weston, J.
\newblock Personalizing dialogue agents: I have a dog, do you have pets too?
\newblock In \emph{Proceedings of the 56th Annual Meeting of the Association
  for Computational Linguistics (Volume 1: Long Papers)}, pp.\  2204--2213,
  Melbourne, Australia, 2018. Association for Computational Linguistics.

\bibitem[Zhang et~al.(2015)Zhang, Lu, and Lapata]{zhang2015top}
Zhang, X., Lu, L., and Lapata, M.
\newblock Top-down tree long short-term memory networks.
\newblock \emph{arXiv preprint arXiv:1511.00060}, 2015.

\bibitem[Zhu et~al.(2018)Zhu, Lu, Zheng, Guo, Zhang, Wang, and
  Yu]{zhu2018texygen}
Zhu, Y., Lu, S., Zheng, L., Guo, J., Zhang, W., Wang, J., and Yu, Y.
\newblock Texygen: A benchmarking platform for text generation models.
\newblock \emph{SIGIR}, 2018.

\end{thebibliography}
